\documentclass{article} 
\usepackage{iclr2027_conference,times}

\usepackage{amsmath,amsfonts,bm}

\def\eqref#1{equation~\ref{#1}}

\def\1{\bm{1}}

\DeclareMathAlphabet{\mathsfit}{\encodingdefault}{\sfdefault}{m}{sl}
\SetMathAlphabet{\mathsfit}{bold}{\encodingdefault}{\sfdefault}{bx}{n}

\newcommand{\E}{\mathbb{E}}

\usepackage{url}

\usepackage{amsmath}
\usepackage{amssymb}
\usepackage{mathtools}
\usepackage{amsthm}
\usepackage{multirow}
\usepackage{algorithm}
\usepackage{booktabs}
\usepackage{subcaption}
\usepackage{wrapfig}
\usepackage{caption}
\usepackage{upquote}
\usepackage{colortbl}

\usepackage{algorithmic}

\usepackage[most,breakable]{tcolorbox}

\definecolor{mydarkblue}{rgb}{0,0.08,0.45} 
\usepackage[colorlinks=true, allcolors=mydarkblue]{hyperref}
\hypersetup{bookmarksnumbered=true, bookmarksopen=true}

\newcommand{\ascrl}{RefineRL}

\title{{\ascrl}: Advancing Competitive Programming and Beyond with Self-Refinement Reinforcement Learning}

\author{
Shaopeng Fu$^{1}$\thanks{Work done during an internship at Microsoft Research.}
\quad
Xingxing Zhang$^{2}$ \quad
Li Dong$^{2}$ \quad
Furu Wei$^{2}$
\\
\\
$^{1}$KAUST
\qquad
$^{2}$Microsoft Research
\\
}

\iclrfinalcopy 
\begin{document}

\maketitle

\lhead{Preprint.}

\begin{abstract}
While large language models (LLMs) have demonstrated strong performance on complex reasoning tasks such as competitive programming (CP), existing methods predominantly focus on single-attempt settings, overlooking their capacity for iterative refinement. 
In this paper, we present \textbf{\ascrl}, a novel approach designed to unleash the self-refinement capabilities of LLMs for CP problem solving.
{\ascrl} introduces two key innovations:
(1)~\textit{Skeptical-Agent}, an iterative self-refinement agent equipped with local execution tools to validate generated solutions against public test cases of CP problems.
This agent always remains skeptical of its own outputs and thereby enforces rigorous self-refinement even when validation suggests correctness.
(2)~A reinforcement learning (RL) solution to incentivize LLMs to self-refine with only standard RLVR data (\textit{i.e.}, problems paired with their verifiable answers). 
Extensive experiments on Qwen3-4B and Qwen3-4B-2507 demonstrate that our method yields substantial gains: after our RL training, these compact 4B models integrated with the Skeptical-Agent not only outperform much larger 32B models but also approach the single-attempt performance of 235B models.
Notably, although our models are trained solely on CP coding data, {\ascrl} also substantially improves their performance on mathematical reasoning benchmarks, suggesting that the learned self-refinement behavior can transfer beyond coding-specific tasks.
These findings highlight self-refinement as a promising direction for scaling LLMs' general reasoning ability.
\end{abstract}

\section{Introduction}

Large language models (LLMs) have demonstrated strong reasoning abilities in math competitions and competitive programming (CP).
Proprietary systems like AlphaGeometry~\citep{trinh2024solving,chervonyi2025gold}, OpenAI~o1~\citep{el2025competitive}, and Gemini~\citep{comanici2025gemini} have proven that reasoning LLMs can become medal winners in the most challenging mind competitions.
Meanwhile, open-source models like DeepSeek-R1~\citep{guo2025deepseek} and Qwen3~\citep{yang2025qwen3} have achieved remarkable performance on rigorous benchmarks such as AIME and LiveCodeBench.
While the success of these models is largely attributed to long chain-of-thought~\citep{wei2022chain} that can consume tens of thousands of tokens during generations, LLMs' potential for \textit{iterative improvement} remains under-explored.
Indeed, prevailing evaluations use \textit{single-attempt} settings, where models are expected to derive correct solutions in a single inference pass.

More recent studies adopt self-correction techniques to enable iterative improvement in mathematical proving~\citep{lin2025goedel,zhou2025solving}.
In each turn, the model uses the Lean-4 prover to check whether the generated proof is incorrect and, if so, incorporates the feedback to revise the proof.
In the field of CP, \citet{shi2024can} and \citet{li2025s} represent the primary open-source efforts leveraging self-correction.
They employ Reflexion~\citep{shinn2023reflexion} to iteratively correct solutions until they pass all public test cases.
However, such self-correction methods generally rely on sparse visible verification signals to trigger improvement, leading to a general \textit{false-positive} issue in reasoning: outputs may satisfy available checks while still failing on hidden cases or more comprehensive evaluation.
In CP, this manifests as solutions passing public tests but failing private tests, as public test cases are only a sparse proxy for the hidden test distribution.
Further, as a purely test-time method, Reflexion depends on the base model's intrinsic iterative improvement capabilities, lacking explicit training to optimize the improvement process itself.

To address these challenges, we introduce \textbf{\ascrl}, an approach designed to unleash the latent reasoning capabilities of LLMs for iterative improvement in competitive programming.
Our framework consists of two components: the \textbf{\textit{Skeptical-Agent}} and \textbf{\textit{Self-Refinement Reinforcement Learning}}.
Concretely, the Skeptical-Agent employs a fine-grained agentic workflow tailored to different judging statuses on public test cases to perform refinement.
Unlike previous approaches that blindly trust public validation, our agent remains skeptical, enforcing rigorous self-refinement even after passing all public test cases.
However, such self-refinement capability either does not arise naturally in LLMs or remains too weak to be effective.
We thus propose Self-Refinement RL to explicitly incentivize this capability using only standard RLVR data (\textit{i.e.}, problems paired with their verifiable answers).
By constructing a dataset solely from the model's skeptical refinement trajectories and using a new \textit{squared-incentive reward} function, we train LLMs to navigate the complex self-refinement process.
Self-Refinement RL thus enables our approach to go beyond static prompt engineering.

We evaluate {\ascrl} on two 4B reasoning LLMs, Qwen3-4B and Qwen3-4B-2507.
Results demonstrate that our method yields substantial gains with high data efficiency: using only 4.5K problems from the HardTest dataset~\citep{he2025hardtests} to construct training samples, we significantly improve the problem-solving capabilities of these compact 4B models.
Notably, after Self-Refinement RL, {\ascrl}-4B and {\ascrl}-4B-2507 equipped with the Skeptical-Agent not only outperform the single-attempt performance of larger 32B models, but also rival that of significantly larger 235B models across multiple competitive programming benchmarks.
Moreover, although our models are trained solely on CP coding data, {\ascrl} also brings substantial improvements on mathematical reasoning benchmarks, suggesting that the learned self-refinement behavior for overcoming ``false-positive reasoning'' can generalize beyond coding tasks.
These findings indicate that self-refinement remains a largely under-explored avenue for scaling LLM reasoning ability, with significant potential for improving general reasoning performance beyond parameter scaling alone.

\section{Related Work}

\textbf{Code generation of LLMs.}
Reasoning LLMs are good at generating program code with long chain-of-thought~(CoT)~\citep{liu2024deepseek,yang2025qwen3,comanici2025gemini}.
For competitive programming, a specific code generation task, \citet{ahmad2025opencodereasoning} show that training LLMs on long CoT data distilled from frontier models improves their ability to solve CP problems.
\citet{he2025hardtests}, \citet{wang2025codecontests+}, and \citet{fu2025klear} find that synthesized high-quality test cases boost RL training with verifiable rewards.
\citet{hu2025quest} and \citet{wu2026x} further suggest using fully synthesized CP problems for training.
Beyond model training, several test-time scaling strategies, such as example test case rejection, longest response selection, clustering, and unittest generation, are designed for CP~\citep{el2025competitive,yu2025z1,samadi2025scaling,ma2026scaling,zhang2026scaling,li2026atgen}.
Our work aims to incentivize a new Skeptical-Agent capability for competitive programming via RL training.

\textbf{Self-correction of LLMs.}
Self-correction is a powerful test-time scaling strategy that empowers LLMs to iteratively critique and improve their responses based on internal or external feedback~\citep{shinn2023reflexion,chen2024teaching,li2025s,madaan2023selfrefine,novikov2025alphaevolve}.
Previous works have demonstrated its effectiveness for coding and mathematical reasoning tasks~\citep{shi2024can,chen2025sets,song2025progco}, and recent studies indicate that LLMs can further internalize this capability through post-training~\citep{chen2024teaching,jain2025multiturn,kumar2025training}.
Traditional approaches often rely on the complete history of multi-turn interactions, leading to excessive context length that limits practicality.
More recent studies in competitive-level math proving~\citep{zhou2025solving,lin2025goedel,stepfunprover2025shang,chen2025seed} suggest decomposing the correction process into multiple single-turn feedback-response pairs, allowing the model to focus on immediate corrections without retaining the full trajectory.
Our paper focuses on designing a self-refinement strategy more robust than self-correction for solving CP problems and on effectively using RL (post-)training to enhance such a capability.


\section{Preliminaries}
\label{sec:preliminaries}

\textbf{Competitive Programming (CP).}
A CP problem can be formalized as a triple $(x, T^{\text{pub}}, T^{\text{pri}})$,
where $x$ is the problem statement,
$T^{\text{pub}} = \{(i^{\text{pub}}_k, o^{\text{pub}}_k)\}_{k=1}^{|T^{\text{pub}}|}$ is the set of \textit{public} test cases,
$(i^{\text{pub}}_k, o^{\text{pub}}_k)$ is the input-output pair of the $k$-th public test case,
$T^{\text{pri}} = \{(i^{\text{pri}}_k, o^{\text{pri}}_k)\}_{k=1}^{|T^{\text{pri}}|}$ is the set of \textit{private} test cases,
and $(i^{\text{pri}}_k, o^{\text{pri}}_k)$ is the input-output pair of the $k$-th private test case.
In a CP contest, the problem statement $x$ and the public test cases $T^{\text{pub}}$ are visible to CP contestants, while the private test cases $T^{\text{pri}}$ are not.
The goal of a CP contestant is to write a code $r_{\text{code}}$ based on $x$ and $T^{\text{pub}}$ that passes $T^{\text{pri}}$.

To determine whether the proposed solution code $r_{\text{code}}$ successfully solves the given problem $(x, T^{\text{pub}}, T^{\text{pri}})$, the contest coordinator uses two oracle functions, \textit{i.e.}, an execution function $f(\cdot,\cdot)$ and a judgment function $g(\cdot,\cdot)$, to evaluate $r_{\text{code}}$ on the private test set $T^{\text{pri}}$.
Concretely, for the $k$-th private test case $(i^{\text{pri}}_k, o^{\text{pri}}_k) \in T^{\text{pri}}$, the execution output and judging status are computed as follows,
\begin{align}
    \hat o^{\text{pri}}_k = f(r_{\text{code}}, i^{\text{pri}}_k), 
    \quad\quad
    j^{\text{pri}}_k = g(\hat o^{\text{pri}}_k, o^{\text{pri}}_k), 
    \label{eq:fn_oracle}
\end{align}
where $\hat o^{\text{pri}}_k$ is the execution output, and $j^{\text{pri}}_k$ is the judging status, indicating whether $r_{\text{code}}$ passes $(i^{\text{pri}}_k, o^{\text{pri}}_k)$ and, if not, the failure reason.
We adopt six judging statuses common in real-world CP contests: ``\textbf{Accepted} (AC)'', ``\textbf{Wrong Answer} (WA)'', ``\textbf{Time Limit Exceeded} (TLE)'', ``\textbf{Memory Limit Exceeded} (MLE)'', ``\textbf{Compile Error} (CE)'', and ``\textbf{Runtime Error} (RE)''.
Among them, ``AC'' means the code passes, while ``WA'', ``TLE'', ``MLE'', ``CE'', and ``RE'' specify the failure reason.

Finally, the proposed solution code $x_{\text{code}}$ is determined to successfully solve the given problem $(x,T^{\text{pub}},T^{\text{pri}})$ if it passes all private test cases of the problem.
In other words, $x_{\text{code}}$ is considered to be \textit{correct} if and only if the summation
$(\sum_{k=1}^{|T^{\text{pri}}|} 1[ j^{\text{pri}}_k = \text{AC} ])$
equals to $|T^{\text{pri}}|$.


\textbf{Solving CP problems with LLMs.}
Let $\pi_\theta$ be a reasoning LLM parameterized by $\theta$.
For a given CP problem $(x,T^{\text{pub}},T^{\text{pri}})$, one can directly prompt the LLM $\pi_\theta$ to solve the problem as follows,
\begin{align}
    (r_{\text{cot}}, r_{\text{sol}}, r_{\text{code}}) \sim \pi_\theta(x),
    \label{eq:pmpt:direct-gen}
\end{align}
where $r_{\text{sol}}$ is the generated CoT, $r_{\text{sol}}$ is the generated text solution, and $r_{\text{code}}$ is the solution code (\textit{e.g.}, Python code) corresponding to the text solution.
In this paper, we focus on generating Python solutions, and the prompt template that we use for Eq.~(\ref{eq:pmpt:direct-gen}) is presented in Figure~\ref{fig:pmpt:direct-gen} in Appendix~\ref{app:prompts}.

\begin{figure*}[t]
    \centering
    \begin{subfigure}{0.4\linewidth}
    \centering
    \includegraphics[width=1.0\linewidth]{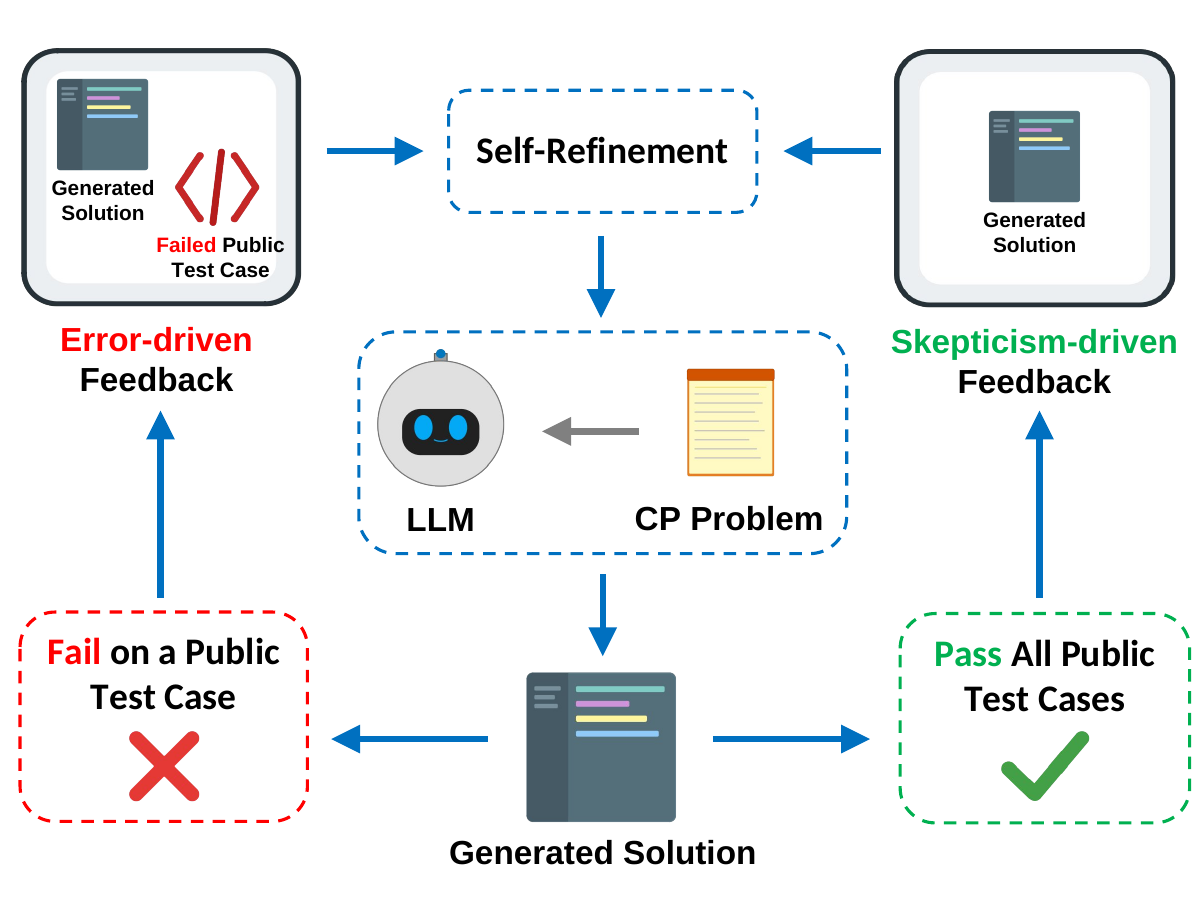}
    \caption{Skeptical-Agent for CP Problems.}
    \end{subfigure}
    \hspace{4em}
    \begin{subfigure}{0.4\linewidth}
    \centering
    \includegraphics[width=1.0\linewidth]{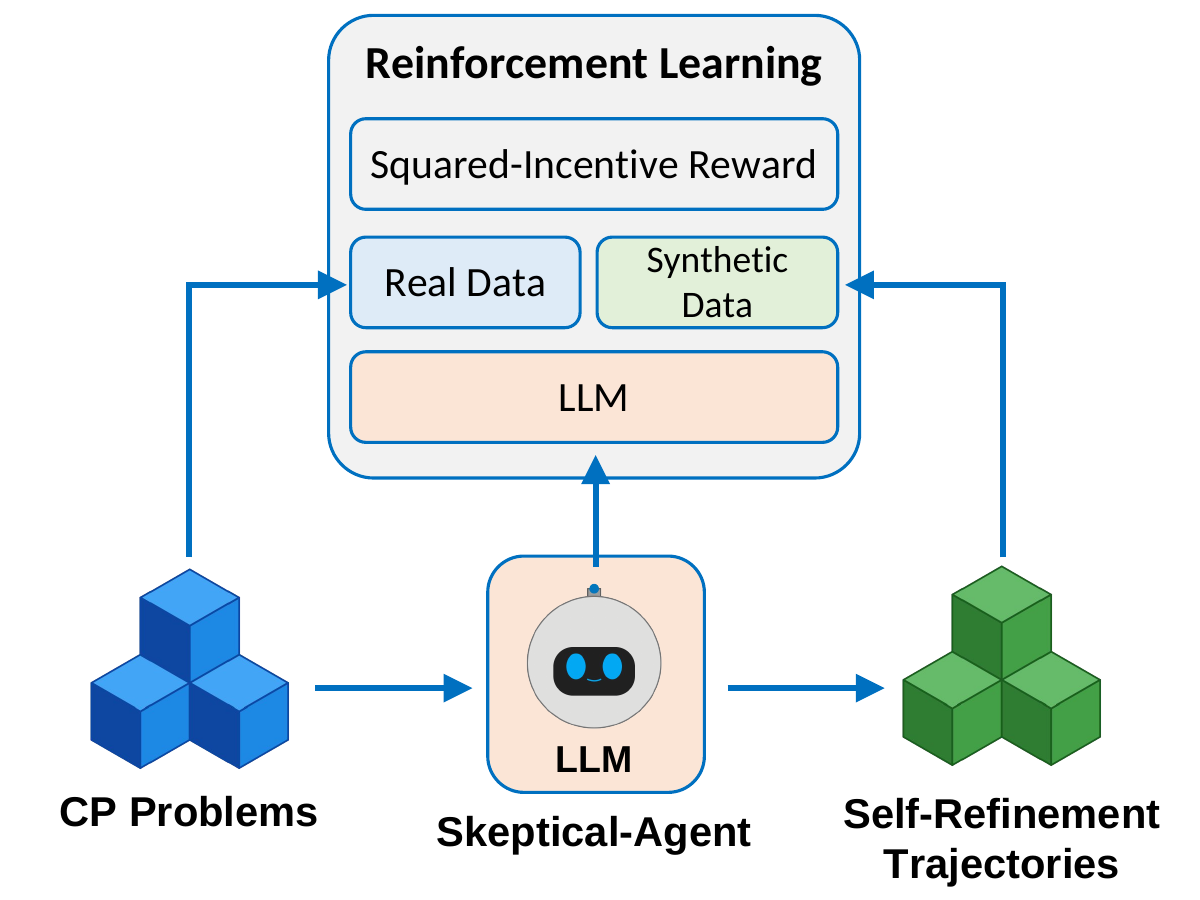}
    \caption{Self-Refinement RL.}
    \end{subfigure}
    \caption{
    Overview of our {\ascrl} approach.
    (a) \textbf{Skeptical-Agent} employs local tools to evaluate generated solutions against visible public test cases and enforces rigorous self-refinement even when evaluation results suggest correctness.
    (b) \textbf{Self-Refinement RL} utilizes self-generated refinement trajectory data from the Skeptical-Agent on real-world CP problems data, along with a novel \textit{Squared-incentive reward} function, to advance the self-refinement capabilities of LLMs.
    }
    \label{fig:method-pipeline}
\end{figure*}

\section{\ascrl}
\label{sec:skeptic}

This section presents our {\ascrl} approach for competitive programming.
We will first motivate the necessity of integrating \textit{skepticism} into iterative improvement-based CP problem solving, and then introduce our proposed \textbf{\textit{Skeptical-Agent}} and \textbf{\textit{Self-Refinement RL}} in detail.

\subsection{Motivation}
\label{sec:skeptic:motivation}

In the domain of CP, \citet{shi2024can} and \citet{li2025s} are primary open-source efforts leveraging self-correction for iterative improvement.
In their approach, the model analyzes the improvement history and corrects its code until the solution passes all visible public test cases or reaches the maximum steps.
However, public test cases are usually sparse and weak, failing to cover the comprehensive distribution of private test cases, particularly edge cases.
This limitation creates a significant risk of \textit{false positives}: a solution passing all public test cases can still fail on unseen private ones.

To tackle false positives, we argue that a promising strategy is to integrate \textit{skepticism} into iterative improvement for competitive programming.
An LLM should not blindly trust evaluations on public test cases but should remain skeptical, continuing refinement even when public evaluations suggest correctness.
Built upon this principle, our Skeptical-Agent utilizes specialized prompts to enforce persistent \textit{self-refinement}, while Self-Refinement RL internalizes this self-refinement capability through post-training, enabling the LLM to discover and resolve defects beyond public test cases.
The high-level designs of Skeptical-Agent and Self-Refinement RL are shown in Figure~\ref{fig:method-pipeline}.

\subsection{Skeptical-Agent}
\label{sec:skeptic:agent}

We now present the detailed design of our Skeptical-Agent.
Its overall pipeline is provided in Algorithm~\ref{algo:test-time-scale} in Appendix~\ref{app:algo}.
Given a CP problem $(x,T^{\text{pub}},T^{\text{pri}})$, Skeptical-Agent proposes and iteratively refines solutions for up to $N$ iterations, using only the problem statement $x$ and the public test cases $T^{\text{pub}}$.
At each iteration, following designs similar to those in \citet{kumar2025training}, \citet{jain2025multiturn}, and \citet{lin2025goedel}, Skeptical-Agent constructs feedback for the next generation solely from the previous attempt.


\textbf{Self-refinement in Skeptical-Agent.}
Let $(r^{(n)}_{\text{cot}}, r^{(n)}_{\text{sol}}, r^{(n)}_{\text{code}})$ be the solution tuple generated in the $n$-th refinement iteration.
To construct feedback $r^{(n)}_{\text{fed}}$ for the $(n+1)$-th generation, we sequentially evaluate the generated code $r^{(n)}_{\text{code}}$ on each public test case from $T^{\text{pub}}$.
For the $k$-th public test case $(i^{\text{pub}}_k, o^{\text{pub}}_k)$, we first obtain an execution output $\hat o^{(n)}_k := f(r^{(n)}_{\text{code}}, i^{\text{pub}}_k)$ (see Eq.~(\ref{eq:fn_oracle})) and then calculate a judging status $j^{(n)}_k := g(\hat o^{(n)}_k, o^{\text{pub}}_k)$, using the oracle execution and judgment functions $f$ and $g$ respectively.
The $(n+1)$-th solution generation process of Skeptical-Agent is formalized as follows,
\begin{align}
    (r^{(n+1)}_{\text{cot}}, r^{(n+1)}_{\text{sol}}, r^{(n+1)}_{\text{code}})
    \sim \pi_\theta(x, r^{(n)}_{\text{fed}}),
    \label{eq:pmpt:augmented-gen}
\end{align}
where $\pi_\theta$ is the reasoning LLM and $r^{(n)}_{\text{fed}}$ is the $n$-th feedback constructed from the evaluation of the $n$-th generated solution on the public test cases set $T^{\text{pub}}$.
The prompt template used by Eq.~(\ref{eq:pmpt:augmented-gen}) is presented in Figure~\ref{fig:pmpt:direct-gen} in Appendix~\ref{app:prompts}.

Then, to maintain skepticism, the Skeptical-Agent will use one of the \textit{error-driven} feedback and \textit{skepticism-driven} feedback during refinement, detailed as follows.

\textbf{Error-driven feedback.}
During the sequential evaluation, if the solution code fails to pass the $k$-th public test case, \textit{i.e.}, if $j^{(n)}_k \in \{\text{WA}, \text{TLE}, \text{MLE}, \text{CE}, \text{RE}\}$, then the evaluation will be terminated and the feedback $r^{(n)}_{\text{fed}}$ obtained from the $n$-th generation is constructed as follows,
\begin{align}
    r^{(n)}_{\text{fed}}
    = \mathcal{P}_{\text{Err}}(r^{(n)}_{\text{sol}}, r^{(n)}_{\text{code}}, i^{\text{pub}}_k, o^{\text{pub}}_k, \hat o^{(n)}_k, j^{(n)}_k),
    \label{eq:aug-fed:p-fail}
\end{align}
where $\mathcal{P}_{\text{Err}}$ is the \textit{error-feedback function} that constructs feedback prompts for different failed judging statuses $j^{(n)}_k$ with different failure reason explanations. The prompt template for $\mathcal{P}_{\text{Err}}$ and the failure reason explanation template are in Figure~\ref{fig:pmpt:p-fail} and Figure~\ref{fig:pmpt:p-fail:explain} in Appendix~\ref{app:prompts}.
This design aims to provide as many failure details as possible to assist the LLM to generate better solutions.

\textbf{Skepticism-driven feedback.}
If the solution code $r^{(n)}_{\text{code}}$ successfully passes all public test cases, \textit{i.e.}, if $j^{(n)}_k = \text{AC}$ for every $(i^{\text{pub}}_k, o^{\text{pub}}_k) \in T^{\text{pub}}$, there might still be bugs or wrong ideas in the solution that can cause the code to fail in edge or extreme private test cases from $T^{\text{pri}}$.
However, in this setting, there is no test case that can be used for the feedback and assisting the next-time solution generation in Skeptical-Agent.
To tackle such a challenge, we propose to instead construct the feedback solely using the generated solution text $r^{(n)}_\text{sol}$ and code $r^{(n)}_{\text{code}}$ as follows,
\begin{align}
    r^{(n)}_{\text{fed}}
    = \mathcal{P}_{\text{Skep}}(r^{(n)}_{\text{sol}}, r^{(n)}_{\text{code}}),
    \label{eq:aug-fed:p-pass}
\end{align}
where $\mathcal{P}_{\text{Skep}}$ is the \textit{Skepticism-feedback function} that constructs prompts asking the LLM to re-check the generated solution $(r^{(n)}_{\text{sol}}, r^{(n)}_{\text{code}})$.
The prompt template for $\mathcal{P}_{\text{Skep}}$ is in Figure~\ref{fig:pmpt:p-pass} in Appendix~\ref{app:prompts}, focusing on checking:
(1)~whether $r^{(n)}_{\text{code}}$ encodes public test cases as short-cuts,
(2)~whether $(r^{(n)}_{\text{sol}}, r^{(n)}_{\text{code}})$ is correct,
and (3)~whether the time and space complexities of $(r^{(n)}_{\text{sol}}, r^{(n)}_{\text{code}})$ are optimal.

\textbf{Overall agent-loop workflow.}
As aforementioned, in the agent-loop workflow, the Skeptical Agent performs up to $N$ self-refinement iterations.
We further adopt the following two additional \textit{rules}, with two additional hyperparameters $M_1$ and $M_2$, to control the workflow:

\begin{itemize}
\item \textbf{Restart Rule.}
If the agent fails to generate a solution that passes all test cases in $T^{\text{pub}}$ for $M_1$ consecutive attempts, then in the $(M_1+1)$-th trial, it will discard the feedback from the previous attempt and generate a solution solely based on the problem statement $x$.

\item \textbf{Termination Rule.}
If the agent generates solutions that passes $T^{\text{pub}}$ for $M_2$ consecutive attempts, it will terminate the refinement and select the last attempt as the final solution.
\end{itemize}

The \textbf{Restart Rule} is designed to prevent the self-refinement process from getting trapped in local optima.
Conversely, the \textbf{Termination Rule} compels the LLM to re-examine and rectify potential latent bugs in the proposed solution, even when it already satisfies all public test cases.

\subsection{Self-Refinement Reinforcement Learning}
\label{sec:skeptic:rl-training}



We now detail our Self-Refinement RL for improving the self-refinement capability of the Skeptical-Agent.
We use the GRPO~\citep{guo2025deepseek} algorithm, excluding KL-regularization, to perform RL training.
The objective function for training an LLM $\pi_\theta$ is formalized as follows,
\begin{align*}
    \mathcal{J}_{\text{GRPO}}(\theta)
    = \E_{q \sim D, \{o_i\}_{i=1}^G \sim \pi_{\theta_{\text{old}}}(\cdot|q)}
    \frac{1}{G} \sum_{i=1}^G \min \Bigl( s_i \hat{A}_i, \text{clip}\left( s_i, 1-\epsilon, 1+\epsilon \right) \hat{A}_i \Bigr),
\end{align*}
where
$q$ is the input sample drawn from the training set $D$,
$\{o_i\}_{i=1}^G$ is the group of responses,
$s_i := \frac{\pi_{\theta}(o_i|q)}{\pi_{\theta_{\text{old}}}(o_i|q)}$ is the importance sampling ratio,
$\hat{A}_i := \frac{R_i - \text{mean}(R)}{\text{std}(R)}$ is the advantage for the $i$-th response, and $\{R_i\}_{i=1}^G$ is the group of rewards for each response.

To help internalize the capability of skeptical self-refinement, our Self-Refinement RL also adopts two unique designs:
(1)~a novel RL training data synthesis paradigm,
and (2)~a new reward function for better RL training on CP problems.


\textbf{RL training data synthesis.}
From the solution generation process formalized in Eq.~(\ref{eq:pmpt:augmented-gen}) in Section~\ref{sec:skeptic:agent}, the key to advancing the Skeptical-Agent is to train LLMs to better perform self-refinement with the feedback $r^{(n)}_{\text{fed}}$ that can be either \textit{error-driven feedback} (see Eq.~(\ref{eq:aug-fed:p-fail})) or \textit{skepticism-driven feedback} (see Eq.~(\ref{eq:aug-fed:p-pass})).
To this end, in our Self-Refinement RL, for a reasoning LLM $\pi_\theta$, we propose to construct, from a normal CP problem dataset, a training dataset that contains prompts combined with failure/skepticism-feedback synthesized solely based on the LLM $\pi_\theta$ itself.

Concretely, suppose $D_{\text{CP}} := \{(x_q, T_q)\}_{q=1}^{|D|}$ is a CP problem dataset where $x_q$ is the statement of the $q$-th problem and $T_q$ is the corresponding test cases set mixing of both public and private test cases.
We use the LLM $\pi_\theta$ to construct two new self-refinement datasets,
$D_{\text{Err}} := \{(x_q, r^{\text{Err}}_{\text{fed},q}, T_q)\}_{q=1}^{|D^{\text{Err}}|}$
and
$D_{\text{Skep}} := \{(x_q, r^{\text{Skep}}_{\text{fed},q}, T_q)\}_{q=1}^{|D^{\text{Skep}}|}$,
where $r^{\text{Err}}_{\text{fed},q}$ is the error-driven feedback from Eq.~(\ref{eq:aug-fed:p-fail}) and $r^{\text{Skep}}_{\text{fed},q}$ is the skepticism-driven feedback from Eq.~(\ref{eq:aug-fed:p-pass}).
The eventual Self-Refinement RL training dataset $D_{\text{RL}}$ is formalized as the union of the original and self-refinement data, \textit{i.e.}, $D_{\text{RL}} := D_{\text{CP}} \cup D_{\text{Err}} \cup D_{\text{Skep}}$.
During training, for each tuple $(x, r_{\text{fed}}, T)$ from the set $D_{\text{RL}}$, we use the prompt template in Figure~\ref{fig:pmpt:direct-gen} to construct the actual prompt for performing the rollout process of RL.

The construction of the two self-refinement datasets, $D_{\text{Err}}$ and $D_{\text{Skep}}$, is detailed as follows:

\textbf{Error-feedback set $D_{\text{Err}}$.}
For each tuple $(x_q,T_q) \in D_{\text{CP}}$ from the original problem set, we first use the LLM $\pi_\theta$ to generate the solution text $r_{\text{sol},q}$ and code $r_{\text{code},q}$ following Eq.~(\ref{eq:pmpt:direct-gen}).
We then evaluate $r_{\text{code},q}$ on $T_q$ using the execution and judging functions $f$ and $g$ in Eq.~(\ref{eq:fn_oracle}).
If $r_{\text{code},q}$ passes all test cases in $T_q$, we exclude this problem from $D_{\text{Err}}$.
Otherwise, for a failed test case $(i_{q,k}, o_{q,k}) \in T_q$, we record the execution output $\hat o_{q,k} := f(r_{\text{code},q}, i_{q,k})$ and judging status $j_{q,k} := g(\hat o_{q,k}, o_{q,k})$.
We then construct the error-driven feedback as
$r^{\text{Err}}_{\text{fed},q} := \mathcal{P}_{\text{Err}}(r_{\text{sol},q}, r_{\text{code},q}, i_{q,k}, o_{q,k}, \hat o_{q,k}, j_{q,k})$
following Eq.~(\ref{eq:aug-fed:p-fail}), and add the tuple $(x_q, r^{\text{Err}}_{\text{fed},q}, T_q)$ to $D_{\text{Err}}$.

\textbf{Skepticism-feedback set $D_{\text{Skep}}$.}
For each tuple $(x_q,T_q) \in D_{\text{CP}}$ from the original problem set, we similarly use the LLM $\pi_\theta$ to generate the solution text $r_{\text{sol},q}$ and code $r_{\text{code},q}$.
Without checking whether $r_{\text{code},q}$ passes $T_q$, we directly construct the skepticism feedback as
$r^{\text{Skep}}_{\text{fed},q} := \mathcal P_{\text{Skep}}(r_{\text{sol},q}, r_{\text{code},q})$
following Eq.~(\ref{eq:aug-fed:p-pass}).
We then add the tuple $(x_q, r^{\text{Skep}}_{\text{fed},q}, T_q)$ to $D_{\text{Skep}}$ with probability $20\%$ if $r_{\text{code},q}$ passes all test cases in $T_q$, and with probability $40\%$ otherwise.\footnote{
The $20\%/40\%$ rates are heuristic.
For failed solutions, skepticism feedback provides no explicit failure signal, making successful correction rare and often leading to weak group-relative rewards in GRPO, as few rollouts receive substantially higher rewards than the others.
We therefore keep only $40\%$ of such samples rather than all of them.
For passed solutions, preserving correctness is easier and may yield uninformative uniformly high rewards, so we keep them at a lower rate of $20\%$.
}



\textbf{Squared-incentive reward.}
Existing studies of RL training on CP problems usually use a \textit{pass-or-fail} reward function where a generated solution code $x_{\text{code}}$ receives a non-trivial reward from a CP problem $(x, T)$ if and only if it passes all test cases from the set $T$~\citep{he2025hardtests,wang2025codecontests+}.
While such a pass-or-fail reward is close to settings of real-world CP contests such as CodeForces or ICPC where contestants need to pass all test cases to get scores from a single problem, we note that it may hinder RL training.
Specifically, for a given CP problem $(x,T)$, if it is so difficult that the LLM being trained cannot generate any correct solution to pass the test case set $T$, then the LLM cannot obtain any non-trivial reward and thus cannot learn anything from $(x,T)$.

To address this technical challenge, our Skeptical RL training adopt a new \textit{\textbf{Squared-incentive Reward}} function that is specifically designed for RL training on CP data.
Given a CP problem $(x,T)$ and a generated solution code $r_{\text{code}}$, the squared-incentive reward for $r_{\text{code}}$ is calculated as follows,
\begin{align*}
    R(r_{\text{code}}, T) :=  (k^* / |T|)^2,
\end{align*}
where $k^*$ is either $|T|$ or one less than the index of the first test case in $T$ that $r_{\text{code}}$ fails to pass.
Our squared-incentive reward has two advantages.
\textbf{Firstly}, it allows the LLM to obtain meaningful learning signals from difficult CP problems even when it cannot produce a fully correct solution during RL.
\textbf{Secondly}, it discourages hacking early test cases and instead incentivizes the LLM to produce the most correct possible solution to pass later, high-reward test cases.


\section{Experiments}
\label{sec:exp}


\subsection{Experimental Setup}
\label{sec:exp:setup}

\textbf{Inference-time strategies.}
For our proposed Skeptical-Agent (see Algorithm~\ref{algo:test-time-scale} in Appendix~\ref{app:algo}), we denote it as \textbf{Skeptical@16} and set its query budget $N$ to 16 and the remaining two hyperparameters $M_1$ and $M_2$ to 4 and 8 respectively.

We also adopt five inference-time baselines:
(1)~\textbf{Pass@1}: Directly querying the LLM once using Eq.~(\ref{eq:pmpt:direct-gen}) and extracting solutions from its response.
(2)~\textbf{Random@16}: Generating 16 responses in parallel and random choose one that passes all public test cases.
(3)~\textbf{LongCoT@16}: Generating 16 responses in parallel and random choose one that passes all public test cases and has the longest CoT length.
(4)~\textbf{RejSamp@16}: Repeatedly querying the LLM up to 16 times until a solution code that can pass all public test cases is generated.
(5)~\textbf{Reflexion@16}~\citep{shi2024can}: Repeatedly querying the LLM up to 16 times with the full historical generation results until a solution code that can pass the problem is generated.
Note that Random@16 and LongCoT@16 are parallel sampling-based baselines, while RejSamp@16 and Reflexion@16 are sequential sampling-based baselines.

\textbf{Training data of RL.}
We select 4.5K CP problems from the HardTest dataset~\citep{he2025hardtests} that are \textit{suitable} for learning self-refinement to form the base problem set $D_{\text{CP}}$.
We then follow the pipeline described in Section~\ref{sec:skeptic:rl-training} to construct the final Self-Refinement RL training data.
We set the maximum generation length to 32,768 for Qwen3-4B and 65,536 for Qwen3-4B-2507, resulting in approximately 4.2K self-refinement trajectories for each model.
See Appendix~\ref{app:exp:dataset} for full details.

\textbf{RL training.}
We perform GRPO-based RL~\citep{guo2025deepseek} on two 4B LLMs, Qwen3-4B and Qwen3-4B-2507~\citep{yang2025qwen3}.
The maximum rollout length is 32,768 tokens for Qwen3-4B and 65,536 for Qwen3-4B-2507.
The resulting models are denoted \textbf{{\ascrl}-4B} and \textbf{{\ascrl}-4B-2507}.
See Appendix~\ref{app:exp:training} for full RL hyperparameters and details of efficient reward calculation.

\textbf{Evaluations.}
We use two versions of LiveCodeBench~\citep{jain2025livecodebench}, \textbf{v5}~(2024.10-2025.01; 167 problems) and \textbf{v6}~(2025.01-2025.03; 175 problems), and the USACO benchmark~\citep{shi2024can}~(307 problems) to evaluate the capability of LLMs in solving CP problems.
We also use three years of AIME datasets (AIME24/25/26) to evaluate the math reasoning capability.
For each LLM, we set the sampling temperature and top-$p$ to 0.6 and 0.95.
For the max generation token length, we set 32,768 for Qwen3-4B and {\ascrl}-4B and 65,536 for other models.
All models are evaluated using our self-implemented evaluation toolkit.
See Appendix~\ref{app:judge-env:eval} for details.


\subsection{Results Analysis}
\label{sec:exp:results}

We present the RL training logs in Appendix~\ref{app:additional-results}.
Here, we focus on analyzing the CP problem-solving performance gains by the Skeptical-Agent and Self-Refinement RL.

\begin{table}[t]
\caption{
Performance on various CP and math benchmarks.
We compare our {\ascrl} models (4B and 4B-2507 versions) against baselines ranging from 4B to 235B parameters.
\textbf{Bold} indicates the best performance, and \underline{underline} indicates the second-best performance.
Our best {\ascrl} model equipped with Skeptical-Agent ({\ascrl}-4B-2507 + Skeptical@16) outperforms most 32B baselines and achieves performance comparable to the 235B model.
}
\label{tab:main-results}
\centering
\tiny
\begin{tabular}{l c c c c c c}
& \multicolumn{3}{c}{\textbf{CP Benchmarks}} & \multicolumn{3}{c}{\textbf{Math Benchmarks}} \\
\toprule
Model & LCB-v5 & LCB-v6 & USACO & AIME24 & AIME25 & AIME26 \\

\midrule
Qwen3-235B-A22B~\citep{yang2025qwen3} & 67.25 & 59.93 & 60.71 & 87.29 & 77.92 & 82.71 \\
Qwen3-235B-A22B-2507~\citep{yang2025qwen3} & 79.08 & 72.50 & 75.51 & 93.33 & 90.63 & 90.00 \\

\midrule
Qwen3-8B~\citep{yang2025qwen3} & 55.39 & 49.32 & 46.17 & 77.92 & 64.58 & 64.58 \\
Dpsk-R1-0528-Qwen3-8B~\citep{guo2025deepseek} & 59.84 & 51.84 & 43.16 & 79.37 & 66.46 & 69.38 \\
X-Coder-RL-Qwen3-8B~\citep{wu2026x} & 55.80 & 48.39 & 43.40 & 33.75 & 26.67 & 33.13 \\
Qwen3-30B-A3B~\citep{yang2025qwen3} & 60.37 & 56.00 & 52.71 & 82.08 & 70.21 & 70.42 \\
Qwen3-30B-A3B-2507~\citep{yang2025qwen3} & \underline{70.55} & 64.61 & 64.68 & \underline{91.46} & \underline{85.21} & \underline{87.33} \\
Qwen3-32B~\citep{yang2025qwen3} & 62.69 & 59.25 & 56.11 & 82.92 & 72.71 & 74.58 \\
OCR-Nemotron-1.1-32B~\citep{ahmad2025opencodereasoning} & 70.47 & 62.25 & 61.97 & 30.21 & 22.08 & 21.46 \\
OR-Nemotron-32B~\citep{ahmad2025opencodereasoning} & 67.03 & \underline{66.07} & \textbf{66.92} & 88.54 & 82.50 & 84.79 \\
Qwen3-4B~\citep{yang2025qwen3} & 52.51 & 48.29 & 40.76 & 72.71 & 63.96 & 65.00 \\
Qwen3-4B-2507~\citep{yang2025qwen3} & 62.65 & 55.21 & 52.44 & 84.38 & 75.63 & 80.83 \\
\rowcolor{green!10}
{\ascrl-4B (Ours)} & 56.40 & 48.79 & 44.75 & 71.67 & 65.21 & 64.79 \\
\rowcolor{green!10}
{\quad + Skeptical@16 (Ours)} & 64.07 & 56.54 & 50.61 & 78.96 & 72.92 & 72.08 \\
\rowcolor{green!10}
{\ascrl-4B-2507 (Ours)} & 70.40 & 61.57 & 59.81 & 87.92 & 77.50 & 82.08 \\
\rowcolor{green!10}
{\quad + Skeptical@16 (Ours)} & \textbf{75.97} & \textbf{67.82} & \underline{64.66} & \textbf{92.08} & \textbf{88.75} & \textbf{88.33} \\

\bottomrule
\end{tabular}

\end{table}

\textbf{CP performance against baseline models.}
We compare the CP performance of our \ascrl-4B and \ascrl-4B-2507 models against strong baselines in Table~\ref{tab:main-results}.
We observe two key findings.
\textbf{Firstly}, although Self-Refinement RL targets self-refinement, it also substantially improves standard Pass@1 performance.
For example, \ascrl-4B achieves performance comparable to 8B models (\textit{e.g.}, Qwen3-8B), while \ascrl-4B-2507 rivals 32B models (\textit{e.g.}, OCR-Nemotron-1.1-32B).
\textbf{Secondly}, with our inference-time Skeptical-Agent strategy (\textit{i.e.}, Skeptical@16), \ascrl-4B-2507 outperforms most 32B baselines and approaches the larger Qwen3-235B-A22B-2507.
Specifically, it reduces the performance gap to Qwen3-235B-A22B-2507 from around 17\% to below 5\% on both LiveCodeBench~v5 and v6, and from 23\% to 11\% on USACO.
These results show that our approach effectively unlocks the self-refinement capability of LLMs, enabling compact 4B models to achieve strong CP performance with efficient training and inference.

\textbf{Generalization to mathematical reasoning.}
We then evaluate whether the self-refinement capability learned from CP data generalizes to mathematical reasoning.
Since there are no executable test cases for math problems, we apply only skepticism-driven refinement ($M_2 = 8$ and $N=16$) in Skeptical-Agent.
As shown in Table~\ref{tab:main-results}, although {\ascrl} models are trained solely on CP coding data, they still achieve clear improvements across three years of AIME benchmarks.
In particular, our best model, {\ascrl}-4B-2507 with Skeptical-Agent, outperforms all 30B/32B baselines and reduces the gap to Qwen3-235B-A22B-2507 from roughly 10--15\% to below 2\%.
These results suggest that our method not only improves CP performance, but also enhances LLMs' general self-refinement capability, helping them overcome false-positive reasoning beyond coding tasks.

\begin{figure}[t]
\centering

\begin{minipage}[t]{0.44\linewidth}
\centering
\captionof{table}{
Performance of inference-time strategies on RefineRL-4B-2507.
Our Skeptical@16 strategy outperforms all compared baselines across all settings.
}
\label{tab:tts-perf:refinerl-4b-2507}
\tiny
\begin{tabular}{l c c c}
\toprule
Model \& Strategy & LCB-v5 & LCB-v6 & USACO \\

\midrule
{{\ascrl}-4B-2507 (Ours)} & 70.40 & 61.57 & 59.81 \\
\quad + Random@16 & 74.06 & 64.14 & 61.39 \\
\quad + LongCoT@16 & 69.65 & 58.82 & 51.97 \\
\quad + RejSamp@16 & 74.25 & 64.71 & 62.32 \\
\quad + Reflexion@16 & 74.36 & 64.29 & 62.77 \\
\rowcolor{green!10}
\quad {+ Skeptical@16 (Ours)} & \textbf{75.97} & \textbf{67.82} & \textbf{64.66} \\

\bottomrule
\end{tabular}
\end{minipage}
\hfill
\begin{minipage}[t]{0.52\linewidth}
\centering
\vspace{0pt}
\includegraphics[width=0.96\linewidth]{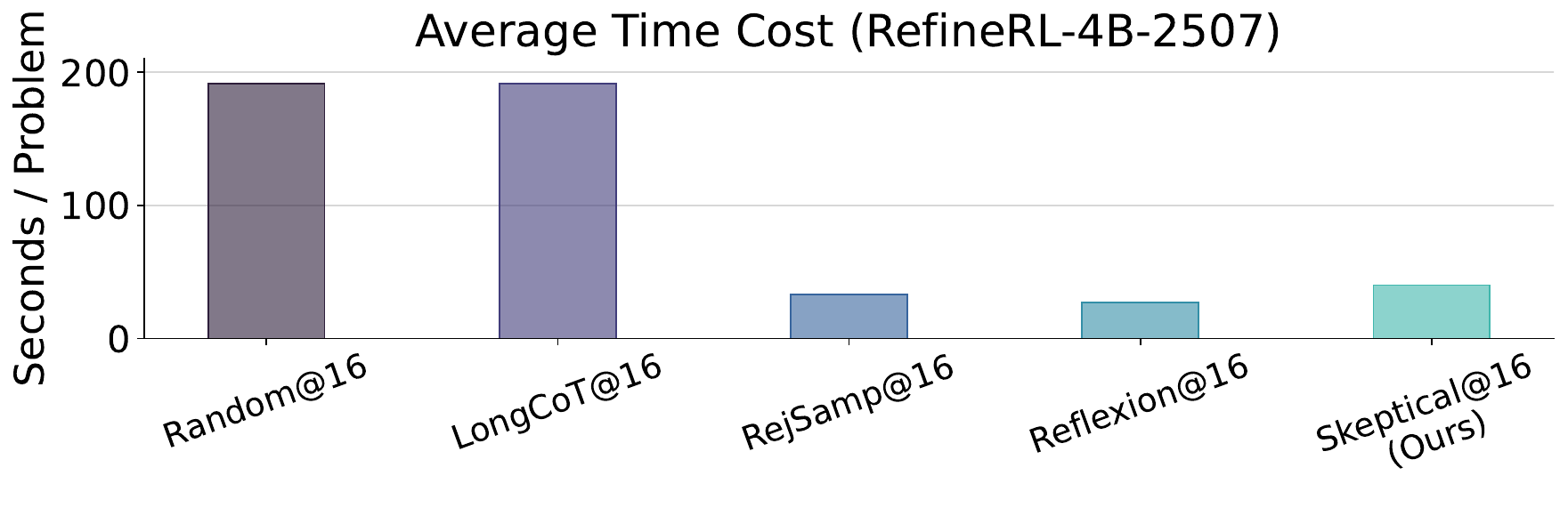}
\caption{
Time cost of inference-time strategies on RefineRL-4B-2507.
Skeptical@16 achieves a speed comparable to other sequential-sampling baselines while being faster than parallel-sampling baselines.
}
\label{fig:tts-time-cost:refinerl-4b-2507}
\end{minipage}
\end{figure}

\textbf{Comparison with inference-time baselines.}
We compare Skeptical-Agent (\textit{i.e.}, Skeptical@16) with various inference-time baselines on RefineRL-4B-2507 in Table~\ref{tab:tts-perf:refinerl-4b-2507}~(performance) and Figure~\ref{fig:tts-time-cost:refinerl-4b-2507}~(time cost).
Similar results on RefineRL-4B are provided in Appendix~\ref{app:tts-results-refinerl-4b}.
Our Skeptical@16 consistently outperforms parallel-sampling baselines (\textit{i.e.}, Random@16 and LongCoT@16) across all three benchmarks.
Notably, compared with Random@16, Skeptical@16 achieves both better performance and about $4\times$ faster inference.
This shows that our gains come from more effective use of the sampling budget, rather than simply increasing parallel test-time compute.

We further compare with other sequential-sampling baselines (\textit{i.e.}, RejSamp@16 and Reflexion@16) to isolate the source of improvement.
Skeptical@16 outperforms RejSamp@16, indicating that the gains do not merely come from stochastic re-sampling until public tests pass.
It also surpasses Reflexion@16, which may prematurely accept false-positive solutions, showing that persistent skepticism helps uncover and fix latent bugs beyond public test cases.


\begin{wraptable}{r}{0.50\textwidth}
\vspace{-1em}
\centering
\caption{
Impact of Self-Refinement RL on reasoning depth.
We compare the average number of self-refinement rounds and the response length per round for models before and after RL training.
}
\label{tab:response-stats}
\tiny
\begin{tabular}{l c c}
\toprule
Model \& Strategy & Avg. Refine Rounds & Avg. Response Len. \\

\midrule
Qwen3-4B & 10.38 & 10,297.97 \\
{\ascrl}-4B (Ours) & 9.78 & 11,371.62 \\

\midrule
Qwen3-4B-2507 & 9.26 & 10,330.52 \\
{\ascrl}-4B-2507 (Ours) & 8.66 & 17,558.65 \\

\bottomrule
\end{tabular}
\end{wraptable}
\textbf{Effect of RL training on model responses.}
We analyze the behavioral changes induced by Self-Refinement RL in Table~\ref{tab:response-stats}.
After RL training, the average number of self-refinement rounds slightly decreases, while the average response length increases substantially.
This suggests that our objective does not simply encourage the model to ``try more times'', but promotes deeper reasoning within each attempt.
By generating longer and more detailed CoT, the model effectively trades the quantity of attempts for reasoning quality.
Crucially, this substantial increase in reasoning length for {\ascrl}-4B-2507 correlates with the superior performance gains reported in our main results, reinforcing the hypothesis that extended, comprehensive CoT is instrumental in solving complex CP problems.

\begin{wraptable}{r}{0.42\linewidth}
\vspace{-1em}
\centering
\caption{
Degradation rates of Skeptical-Agent (Skeptical@16) on LCB-v6.
}
\label{tab:skep-degradation}
\tiny
\begin{tabular}{l c c}
\toprule
Model (w/ Skeptical@16) & (\%) / Round & (\%) / Problem \\
\midrule
RefineRL-4B & 1.13 & 1.86 \\
RefineRL-4B-2507 & 0.53 & 2.29 \\
\bottomrule
\end{tabular}
\end{wraptable}
\textbf{Degradation analysis of Skeptical-Agent.}
We examine whether skepticism-driven refinement may hurt correct solutions.
Table~\ref{tab:skep-degradation} reports two degradation metrics on LiveCodeBench-v6: \textit{round-level degradation rate}, measuring how often a correct solution becomes incorrect after one refinement round, and \textit{problem-level degradation rate}, measuring the fraction of problems degraded by over-refinement.
We find that degradation is rare for both models, with round-level rates below $1.2\%$ and problem-level rates below $2.3\%$.
This suggests that over-refinement is not a major practical issue for Skeptical-Agent.

\subsection{Ablation Studies}
\label{sec:exp:ablation}

\begin{figure}[t]
\centering

\begin{minipage}[t]{0.4\linewidth}
\centering

\captionof{table}{
Performance of Qwen3-4B-2507 on trained with and without self-refinement trajectory data.
}
\label{tab:ablation:self-refinement-data}
\tiny
\begin{tabular}{l c c}
\toprule
Model \& Strategy & LCB-v5 & LCB-v6 \\

\midrule
Qwen3-4B-2507 & 62.65 & 55.21 \\
\ + Standard RL & 66.95 & 57.96 \\
\ {+ Self-Refinement RL (Ours)} & \textbf{70.40} & \textbf{61.57} \\

\midrule
Qwen3-4B-2507 + Skeptical@16 & 71.29 & 62.07 \\
\ + Standard RL & 72.90 & 63.79 \\
\ {+ Self-Refinement RL (Ours)} & \textbf{75.97} & \textbf{67.82} \\

\bottomrule
\end{tabular}

\end{minipage}
\hfill
\begin{minipage}[t]{0.56\linewidth}
\captionsetup{type=figure}
\centering
\vspace{0pt}

    \begin{minipage}[t]{0.30\linewidth}
        \centering
        \includegraphics[width=\linewidth]{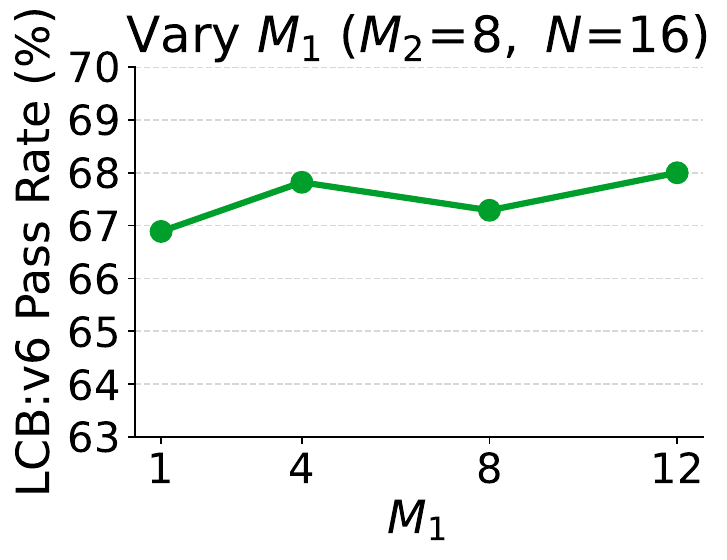}
        \subcaption{Results of $M_1$.}
    \end{minipage}
    \hfill
    \begin{minipage}[t]{0.30\linewidth}
        \centering
        \includegraphics[width=\linewidth]{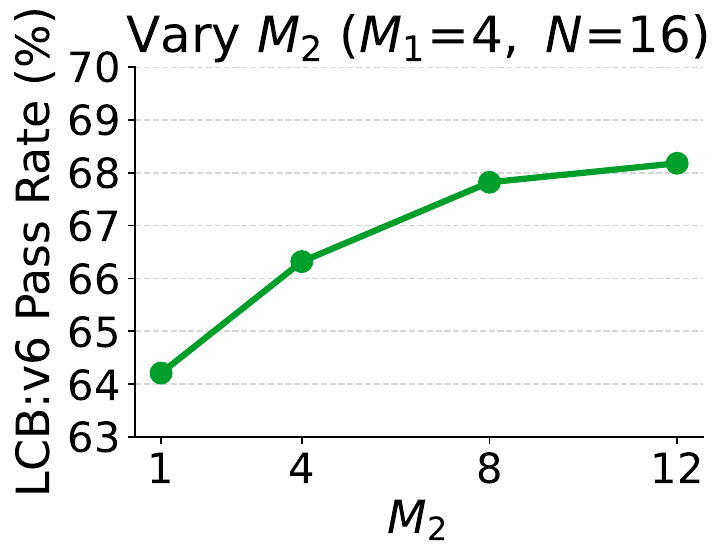}
        \subcaption{Results of $M_2$.}
        \label{fig:ablation-skep-hparams:m2}
    \end{minipage}
    \hfill
    \begin{minipage}[t]{0.30\linewidth}
        \centering
        \includegraphics[width=\linewidth]{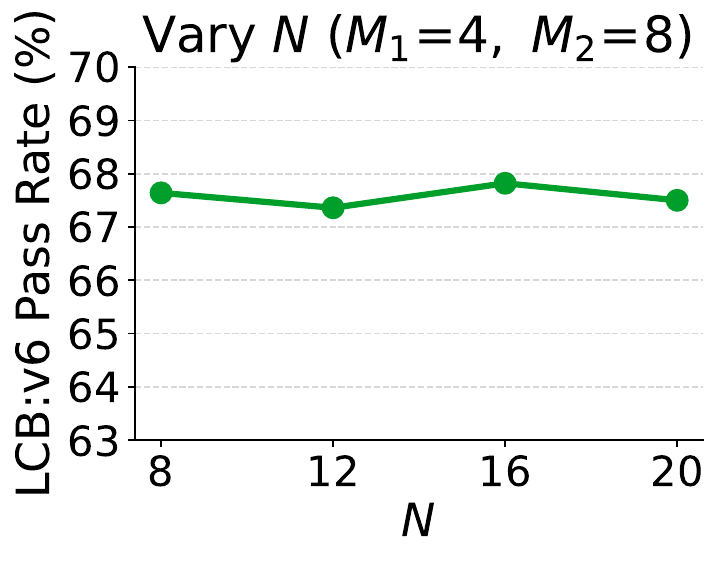}
        \subcaption{Results of $N$.}
    \end{minipage}
    \caption{
    Impact of $(N,M_1,M_2)$ of Skeptical-Agent on RefineRL-4B-2507 and LCB-v6.
    Notably, $M_2=1$ disables skepticism-driven refinement.
    }
\label{fig:ablation-skep-hparams}

\end{minipage}

\end{figure}

\textbf{Hyperparameters of Skeptical-Agent.}
We study how the three key hyperparameters of Skeptical-Agent, \textit{i.e.}, $M_1$, $M_2$, and $N$, affect its performance.
Figure~\ref{fig:ablation-skep-hparams} reports the results on RefineRL-4B-2507 and LCB-v6.
We find that Skeptical-Agent is relatively stable with respect to $M_1$ and $N$.
In contrast, it is more sensitive to $M_2$: increasing $M_2$ from $1$ to $8$ substantially improves the pass rate (see Figure~\ref{fig:ablation-skep-hparams:m2}).
Recall that $M_2$ controls the degree of skepticism-driven refinement, and $M_2=1$ disables it.
This confirms that enforcing rigorous self-refinement is a key step of Skeptical-Agent.

\begin{table}[t]
\centering

\begin{minipage}[t]{0.53\linewidth}
\centering
\caption{
Performance of models trained via Skeptical-RL with different reward designs.
}
\label{tab:ablation:reward}
\tiny
\begin{tabular}{l c c c c}
\toprule
\multirow{3}{10em}{Qwen3-4B-2507\\Self-Refinement RL} & \multicolumn{2}{c}{No-strategy} & \multicolumn{2}{c}{Skeptical@16} \\
\cmidrule(lr){2-3}
\cmidrule(lr){4-5}
& LCB-v5 & LCB-v6 & LCB-v5 & LCB-v6 \\

\midrule
+ Pass-or-fail Reward & 64.22 & 55.36 & 70.55 & 61.43 \\
+ Linear Reward & 68.23 & 59.86 & 74.03 & 65.07 \\
+ Squared-incentive Reward & \textbf{70.40} & \textbf{61.57} & \textbf{75.97} & \textbf{67.82} \\

\bottomrule
\end{tabular}
\end{minipage}
\hfill
\begin{minipage}[t]{0.44\linewidth}
\centering
\caption{
Ablation on model scale.
RefineRL-8B is trained with the same setting as RefineRL-4B.
}
\label{tab:ablation-8b-rl}
\tiny
\begin{tabular}{l c c}
\toprule
Model & LCB-v5 & LCB-v6 \\
\midrule
Qwen3-8B & 55.39 & 49.32 \\
RefineRL-8B & 57.67 & 51.79 \\
RefineRL-8B + Skeptical@16 & \textbf{66.02} & \textbf{59.39} \\
\bottomrule
\end{tabular}
\end{minipage}
\end{table}

\textbf{Effect of self-refinement trajectory data.}
We then analyze whether the gains of Self-Refinement~RL mainly come from self-generated refinement trajectories.
As a baseline, we train ``Standard~RL'' models only on the original 4.5K HardTest CP problems, using the same squared-incentive reward for a fair comparison.
Table~\ref{tab:ablation:self-refinement-data} shows that models trained with self-refinement trajectories consistently outperform Standard RL baselines, both under Skeptical-Agent and in the single-attempt Pass@1 setting.
This suggests that refinement trajectories help the model internalize debugging behavior, leading to stronger reasoning than training on problem statements alone.

\textbf{Effect of reward functions.}
We next study the efficacy of our squared-incentive reward by comparing it with two alternatives: a binary pass-or-fail reward and a linear reward (see Appendix~\ref{app:exp:reward-fns} for reward definitions), while keeping the self-refinement trajectory data fixed.
As shown in Table~\ref{tab:ablation:reward}, the binary reward yields only weak improvements, while the linear reward performs better with denser feedback.
However, both are consistently worse than our squared-incentive reward across all settings.
This suggests that dense feedback alone is insufficient: by down-weighting easy gains on early test cases and amplifying rewards for later ones, the squared-incentive reward better discourages shallow test-case hacking and provides stronger guidance for complex CP tasks.

\textbf{Ablation on model scale.}
We further examine whether our method remains effective beyond 4B models.
To this end, we train RefineRL-8B from Qwen3-8B using the same pipeline and hyperparameters as RefineRL-4B.
As shown in Table~\ref{tab:ablation-8b-rl}, RefineRL-8B + Skeptical@16 improves Qwen3-8B from 55.39\% to 66.02\% on LCB-v5 and from 49.32\% to 59.39\% on LCB-v6.
These consistent gains show that the benefits of Self-Refinement RL and Skeptical-Agent are not limited to 4B models, but can also scale to stronger 8B backbones.

\section{Conclusion}


We present {\ascrl}, an approach to incentivize LLM self-refinement in competitive programming.
To mitigate unreliable public tests, we introduce \textit{Skeptical-Agent}, which maintains skepticism toward generated solutions and enforces refinement even when local evaluations suggest correctness.
We further propose \textit{Self-Refinement RL}, which internalizes this ``doubt and debug'' behavior through self-generated refinement trajectories and a squared-incentive reward.
Empirically, {\ascrl} enables compact 4B models to outperform 32B baselines and approach much larger 235B models on CP benchmarks.
Although trained solely on CP coding data, {\ascrl} also improves math performance, suggesting that self-refinement generalizes to math reasoning.
These results suggest self-refinement as a promising, data-efficient path toward stronger reasoning beyond parameter scaling.

%

\bibliography{cp_rl}
\bibliographystyle{iclr2027_conference}

\newpage
\appendix

\section{Additional Implementation Details}

This section provides additional implementation details omitted from the main text.

\subsection{Algorithm of Skeptical-Agent}
\label{app:algo}

We provide the complete procedure of Skeptical-Agent (see Section~\ref{sec:skeptic:agent}) in Algorithm~\ref{algo:test-time-scale}.

\begin{algorithm}[h]
\caption{Skeptical-Agent for CP Problem Solving}
\label{algo:test-time-scale}
\begin{algorithmic}[1]
\REQUIRE
Reasoning LLM $\pi_\theta$,
problem statement $x$,
public test cases set $T^{\text{pub}}$,
hyperparameters $(N, M_1, M_2)$,
oracle execution and judging functions $f$ and $g$.
\ENSURE Optimal solution text $r^{(*)}_{\text{sol}}$ and code $r^{(*)}_{\text{code}}$.
    \STATE $c_1 \leftarrow 0$, \ $c_2 \leftarrow 0$ \hfill $\triangleright \ \text{Initialize temporary counters}$
    \STATE $r^{(*)}_{\text{sol}}, r^{(*)}_{\text{code}}, r^{(0)}_{\text{fed}} \leftarrow \text{Null}$ \hfill $\triangleright \ \text{Initialize as empty value}$
    \FOR{$n$ \textbf{in} $1,\cdots, N$}
        \STATE $r^{(n)}_{\text{cot}}, r^{(n)}_{\text{sol}}, r^{(n)}_{\text{code}} \sim \pi_{\theta}(x, r^{(n-1)}_{\text{fed}})$ \hfill $\triangleright \ \text{Eq.~(\ref{eq:pmpt:augmented-gen})}$
        \STATE $(r^{(*)}_{\text{sol}}, r^{(*)}_{\text{code}}) \leftarrow (r^{(n)}_{\text{sol}}, r^{(n)}_{\text{code}})$ \hfill $\triangleright \ \text{Update final solution}$
        \STATE $i_*, o_*, \hat o_*, j_* \leftarrow \text{Null}$ \hfill $\triangleright \ \text{Store judging result}$
        \FOR{$(i^{\text{pub}}_k, o^{\text{pub}}_k)$ \textbf{in} $T^{\text{pub}}$}
            \STATE $\hat o^{(n)}_{k} \leftarrow f(r^{(n)}_{\text{code}}, i^{\text{pub}}_k)$, \ $j^{(n)}_k \leftarrow g(\hat o^{(n)}_k, o^{\text{pub}}_k)$
            \IF{$j^{(n)}_k \ne \text{AC}$}
                \STATE $(i_*, o_*, \hat o_*, j_*) \leftarrow (i^{\text{pub}}_k, o^{\text{pub}}_k, \hat o^{(n)}_k, j^{(n)}_k)$
                \STATE \textbf{break}
            \ENDIF
        \ENDFOR
        \IF{$j_* \ne \text{Null}$}
            \STATE $c_1 \leftarrow ((c_1+1)\mod M_1)$, \ $c_2 \leftarrow 0$
        \ELSE
            \STATE $c_1 \leftarrow 0$, \ $c_2 \leftarrow (c_2+1)$
        \ENDIF
        \IF{$c_2 = M_2$}
            \STATE \textbf{break}
        \ENDIF
        \IF{$c_1 > 0$}
            \STATE $r^{(n)}_{\text{fed}} \leftarrow \mathcal{P}_{\text{Err}}(r^{(n)}_{\text{sol}}, r^{(n)}_{\text{code}}, i_*, o_*, \hat o_*, j_*)$ \hfill $\triangleright \ \text{Eq.~(\ref{eq:aug-fed:p-fail})}$
        \ELSIF{$c_2 > 0$}
            \STATE $r^{(n)}_{\text{fed}} \leftarrow \mathcal{P}_{\text{Skep}}(r^{(n)}_{\text{sol}}, r^{(n)}_{\text{code}})$ \hfill $\triangleright \ \text{Eq.~(\ref{eq:aug-fed:p-pass})}$
        \ELSE
            \STATE $r^{(n)}_{\text{fed}} \leftarrow \text{Null}$
        \ENDIF
    \ENDFOR
    \STATE \textbf{return} $(r^{(*)}_{\text{sol}}, r^{(*)}_{\text{code}})$
\end{algorithmic}
\end{algorithm}

\subsection{Training Data for Self-Refinement RL}
\label{app:exp:dataset}

We construct the training dataset for Skeptical RL based on a subset of the HardTest dataset~\citep{he2025hardtests}, consisting of CP problems that are \textbf{suitable for learning self-refinement}.
Formally, a problem is considered to be \textit{suitable} if it contains at least 8 test cases with both input and output lengths not exceeding 512 tokens.
The idea behind this is that short input-output pairs are essential for both humans and AI models to effectively diagnose errors and facilitate debugging.

To form the base CP problem dataset $D_\text{CP}$, we select suitable problems from the first 1/8 of HardTest.
For each selected problem, we retain all test cases shorter than 512 tokens and randomly sample up to 8 additional test cases from the remainder.
This process results in a base set $D_{\text{CP}}$ consisting of approximately 4.5K CP problems.

Finally, we employ the pipeline described in Section~\ref{sec:skeptic:rl-training} to synthesize the final Self-Refinement RL training data.
We focus on generating Python code, and set the maximum generation length to 32,768 for Qwen3-4B and 65,536 for Qwen3-4B-2507.
For each model, this process generates approximately 4.2K self-refinement trajectories.
Combining these with the base problems yields a total of roughly 8.7K training samples per model.
Detailed statistics of the final training datasets are presented in Table~\ref{tab:dataset-stats}.

\begin{table}[h]
\centering
\caption{
Statistics of the training dataset for Self-Refinement RL.
This table details the sample counts for each interaction category.
``WA/RE/TLE/CE/MLE'' represent trajectories triggered by error-driven feedback.
``AC $\rightarrow$ WA/RE/TLE/CE/MLE'' denote trajectories involving skepticism-driven feedback, where the solution passed public test cases but failed on private ones due to the indicated errors.
}
\label{tab:dataset-stats}
\small
\begin{tabular}{lrr}
\toprule
\textbf{Data Type} & \textbf{Qwen3-4B} & \textbf{Qwen3-4B-2507} \\
\midrule
\textit{Initial CP Problems from $D_{\text{CP}}$} & 4,500 & 4,500 \\
\midrule
\textit{Error-driven Trajectory} & 2,811 & 2,721 \\
\quad [True Negative] Wrong Answer (WA) & 2,127 & 2,077 \\
\quad [True Negative] Runtime Error (RE) & 377 & 364 \\
\quad [True Negative] Time Limit Exceeded (TLE) & 307 & 280 \\
\quad [True Negative] Compile Error (CE) & 0 & 0 \\
\quad [True Negative] Memory Limit Exceeded (MLE) & 0 & 0 \\
\midrule
\textit{Skepticism-driven Trajectory} & 1,449 & 1,471 \\
\quad [True Positive] Accepted (AC) & 309 & 370 \\
\quad [False Positive] AC $\rightarrow$ WA & 863 & 856 \\
\quad [False Positive] AC $\rightarrow$ RE & 148 & 139 \\
\quad [False Positive] AC $\rightarrow$ TLE & 129 & 106 \\
\quad [False Positive] AC $\rightarrow$ CE & 0 & 0 \\
\quad [False Positive] AC $\rightarrow$ MLE & 0 & 0 \\
\midrule
\textbf{Total Number} & \textbf{8,760} & \textbf{8,692} \\
\bottomrule
\end{tabular}
\end{table}

\subsection{Reward Function Details}
\label{app:exp:reward-fns}

This section provides the definitions of the reward functions used in our ablation study in Section~\ref{sec:exp:ablation}.
Given a CP problem $(x,T)$ and a generated solution code $r_{\text{code}}$, let $k^*$ denote either $|T|$ if $r_{\text{code}}$ passes all test cases, or one less than the index of the first test case in $T$ that $r_{\text{code}}$ fails to pass.
We compare the following three reward functions.

\textbf{Pass-or-fail reward.}
This is the standard sparse reward commonly used in code RL:
\begin{align*}
    R_{\text{pass-or-fail}}(r_{\text{code}},T)
    :=
    1\left[k^* = |T|\right],
\end{align*}
where $1[\cdot]$ is an indicator function.
It gives reward $1$ only when the solution passes all test cases, and $0$ otherwise.

\textbf{Linear reward.}
The linear reward gives partial credit according to the fraction of test cases passed before the first failure:
\begin{align*}
    R_{\text{linear}}(r_{\text{code}},T)
    :=
    k^* / |T|.
\end{align*}
Compared with the pass-or-fail reward, it provides denser feedback for partially correct solutions.

\textbf{Squared-incentive reward (ours).}
Our proposed reward squares the fraction of passed test cases:
\begin{align*}
    R(r_{\text{code}},T)
    :=
    \left(k^* / |T| \right)^2.
\end{align*}
This design down-weights easy gains from early test cases while amplifying the reward for passing later ones, encouraging the model to solve the problem more completely.

\subsection{RL Training Details}
\label{app:exp:training}

We use the VeRL framework~\footnote{\url{https://github.com/volcengine/verl}}~\citep{sheng2024hybridflow} to perform RL training via the GRPO~\citep{guo2025deepseek} algorithm, where the KL-regularization term is excluded.
both the batch size and mini-batch size are set to 32 and
For the rollout hyperparameters of GRPO, we set the rollout number to 8, the maximum response length to 32,768 for Qwen3-4B and 65,536 for Qwen3-4B-2507, the temperature to 1.0, and the top-$p$ to 1.0.
AdamW is adopted as the optimizer, where the learning rate is fixed to $10^{-6}$ throughout the RL training.
The two RL trained models from Qwen3-4B and Qwen3-4B-2507 are denoted as \textbf{{\ascrl}-4B} and \textbf{{\ascrl}-4B-2507}, respectively.
Each RL training experiment is conducted on 32 NVIDIA A100@80G GPUs.

Calculating rewards for CP data requires executing LLM-generated code against multiple private test cases, which is a CPU-intensive process.
To enable efficient and scalable reward computation during RL training, we integrate our evaluation logic with the VeRL framework, which is built upon the distributed computing engine Ray~\footnote{\url{https://github.com/ray-project/ray}}.
We leverage Ray to schedule distributed code evaluations, assigning isolated CPU cores for individual execution tasks.
To strictly manage resource consumption during training, we employ \texttt{timeout}~\footnote{\url{https://man.archlinux.org/man/timeout.1}} and \texttt{prlimit}~\footnote{\url{https://man.archlinux.org/man/prlimit.1}} to enforce time and memory constraints, set to 8 seconds and 512MB, respectively.
Finally, to ensure reasoning integrity, we enforce complete CoT generation: only responses containing the CoT closing delimiter (\textit{i.e.}, the \texttt{</think>} tag) are submitted for reward calculation; otherwise, a reward of 0 is assigned.

\subsection{Evaluation Details}
\label{app:judge-env:eval}

Existing evaluations on the LiveCodeBench benchmark typically rely on the official LCB evaluation toolkit~\footnote{\url{https://github.com/LiveCodeBench/LiveCodeBench}} or the Hugging Face LightEval package~\footnote{\url{https://github.com/huggingface/lighteval}}.
However, we argue that these packages are inefficient for large-scale CP solution judging, as they lack native support for high-throughput parallel inference and fail to fully leverage multi-core CPUs for computationally intensive solution code execution.
To address these limitations, we implemented a custom, scalable toolkit for efficient CP performance evaluation.

Our implementation decouples the evaluation pipeline into two distinct stages to ensure robustness and efficiency:
\begin{itemize}
\item
\textbf{Step 1: LLM solution generation.}
We utilize the SGLang inference engine~\footnote{\url{https://github.com/sgl-project/sglang}} to perform high-throughput parallel LLM generation.
We adopt the standard LiveCodeBench solution generation prompts from the LightEval package for LLMs to generate Python solutions, with the exception of our Skeptical-Agent, which incorporates the additional feedback prompts detailed in Appendix~\ref{app:prompts} to perform self-refinement.

\item
\textbf{Step 2: Concurrent code evaluation.}
Consistent with the training setup described in Appendix~\ref{app:exp:training}, we employ the Unix CLI tools \texttt{timeout} and \texttt{prlimit} to enforce strict time and memory constraints.
To maximize resource utilization, we leverage native multiprocessing for the Python's \texttt{asyncio}~\footnote{\url{https://docs.python.org/3/library/asyncio.html}} package to evaluate multiple solutions concurrently.
\end{itemize}

Regarding evaluation settings, for every CP problem, we set a time limit of 6s and a memory limit of 512MB.
All code executions are performed on a 2.5GHz multi-core CPU.
To ensure a fair comparison, all models in our experiments are evaluated using our custom toolkit, with results averaged over 16 runs to ensure stability.
The LLM sampling parameters are set to a temperature of 0.6 and a top-$p$ of 0.95.
The maximum generation length is set to 32,768 tokens for Qwen3-4B and our {\ascrl}-4B, and 65,536 tokens for all other models.
Furthermore, given that the evaluated models are reasoning-oriented, we strictly enforce complete Chain-of-Thought (CoT) integrity; specifically, only responses containing a detected CoT completion tag are submitted for evaluation; otherwise, they are assigned a score of zero.

Additionally, for both CP and math evaluations, each evaluation was conducted on a single node with 8 NVIDIA A100 GPUs.

\clearpage
\section{Additional Experimental Results}
\label{app:additional-results}

This section presents additional experimental results omitted from the main text.

\subsection{Training Dynamics}

We provide a detailed visualization of the training dynamics for our Skeptical RL models and other RL baselines.
Table~\ref{tab:rl_comparison} summarizes the configurations for each strategy, detailing the dataset size and reward functions used for the Qwen3-4B and Qwen3-4B-2507 base models.

\begin{table}[h]
\centering
\caption{
Comparison of RL Training Strategies.
We contrast three RL setups: Vanilla RL (direct generation), Self-Refinement RL with pass-or-fail rewards, and our Self-Refinement RL (with squared-incentive rewards).
Vanilla RL uses only the initial generation subset ($\sim$4.5k), while refinement methods use the full trajectory dataset ($\sim$8.7k).
Our method employs a squared reward to incentivize robust self-verification.
}
\label{tab:rl_comparison}
\footnotesize
\begin{tabular}{l l l l}
\toprule
\textbf{Method} & \textbf{Data Size} & \textbf{Trajectory Type} & \textbf{Reward Function} \\
\midrule
\textit{Vanilla RL} & $\sim$4.5K & None & Squared-incentive \\
\textit{Refinement (Binary)} & $\sim$8.7K (4.5K Real Data + 4.2K Syn. Data) & Refinement & Pass-or-fail \\
\textbf{\ascrl{} (Ours)} & $\sim$8.7K (4.5K Real Data + 4.2K Syn. Data) & Refinement & Squared-incentive \\
\bottomrule
\end{tabular}
\end{table}

\begin{figure}[h]
    \centering
    \begin{subfigure}{0.48\linewidth}
    \centering
    \includegraphics[width=1.0\linewidth]{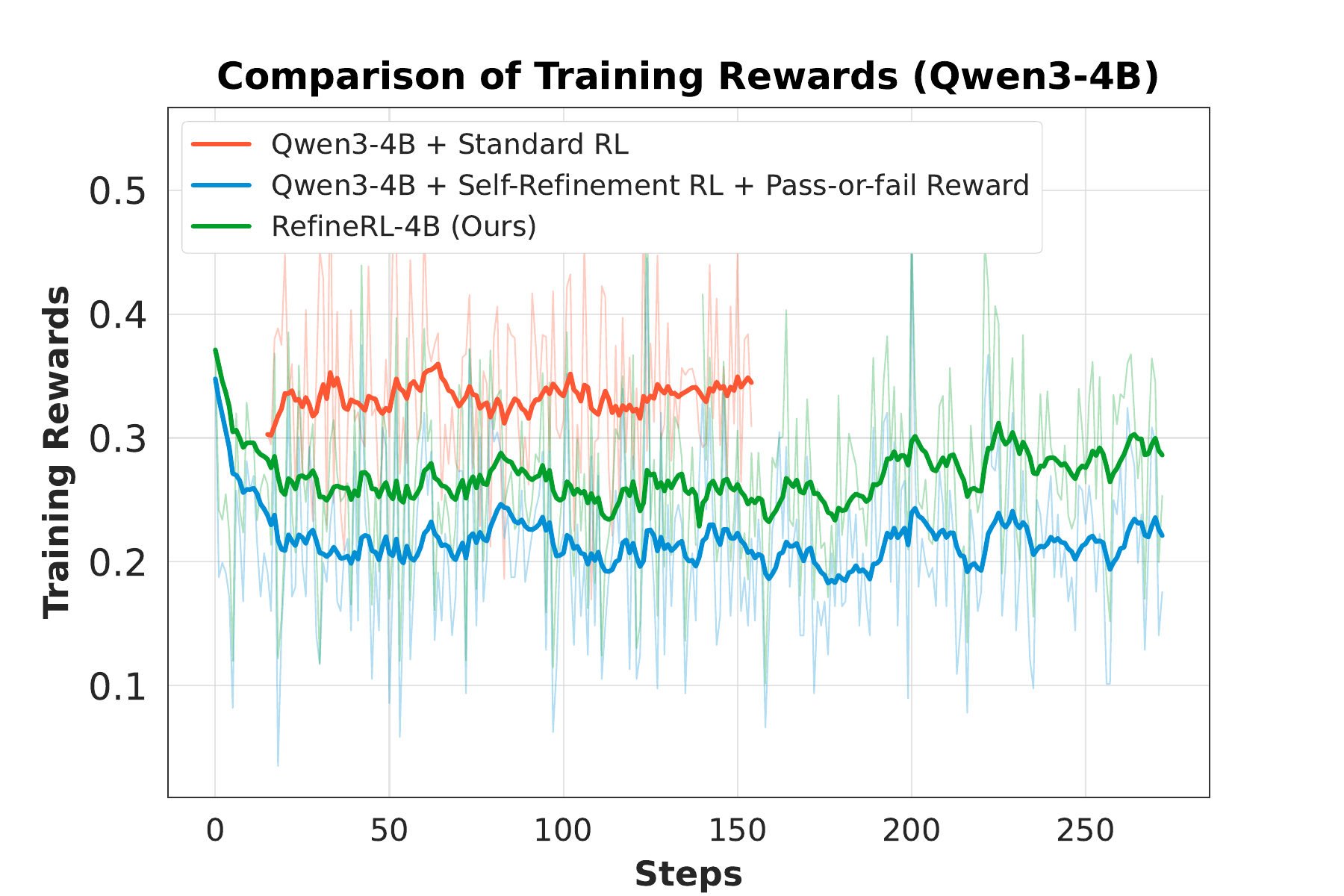}
    \caption{Average reward per step.}
    \end{subfigure}
    \hspace{1em}
    \begin{subfigure}{0.48\linewidth}
    \centering
    \includegraphics[width=1.0\linewidth]{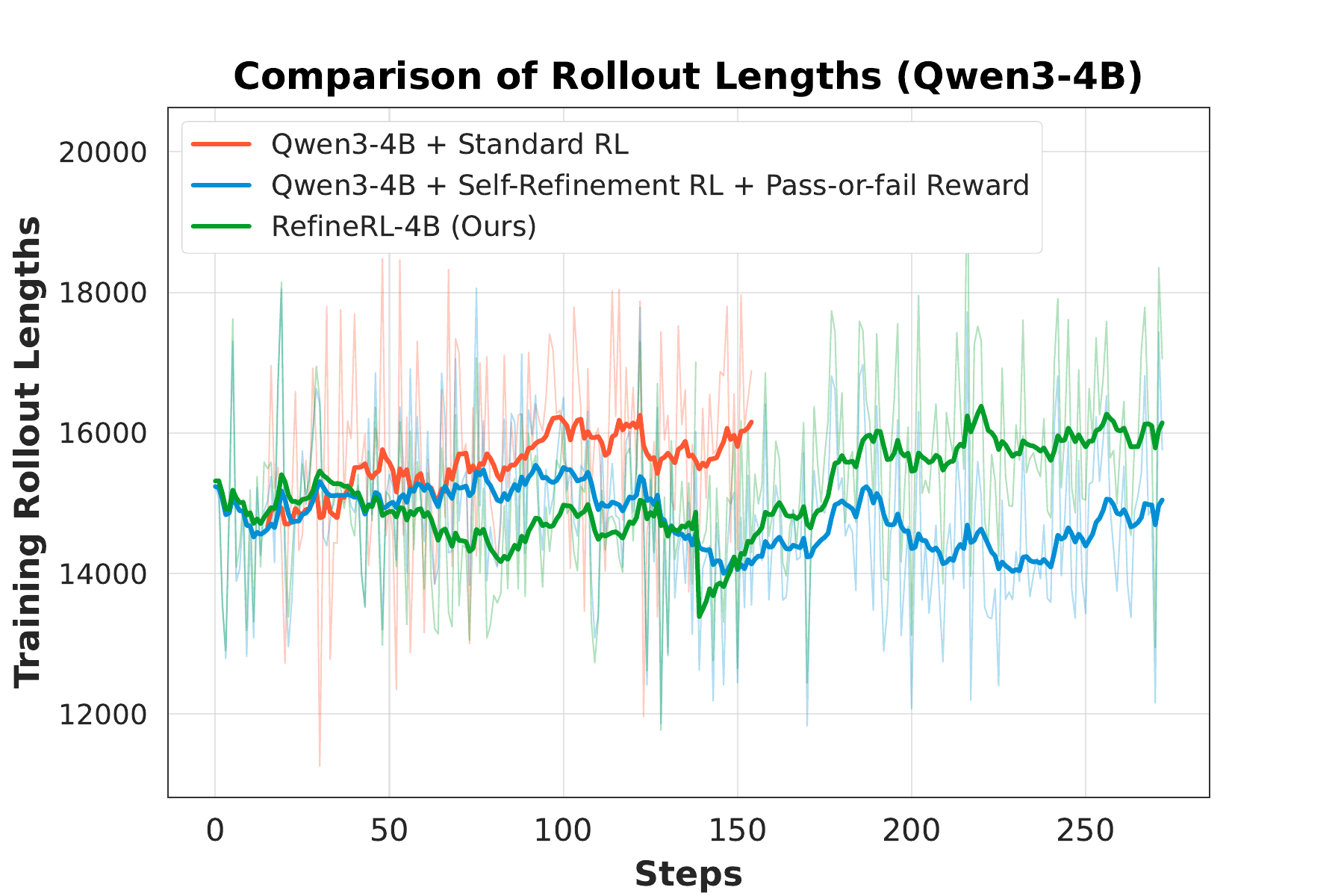}
    \caption{Average rollout length (tokens) per step.}
    \end{subfigure}
    \caption{
    {Training dynamics of Qwen3-4B.}
    The left plot shows the steady increase in reward, while the right plot shows the corresponding growth in response length, reflecting the model's internalization of the skeptical reasoning process.
    }
    \label{fig:curve:qwen3-4b}
\end{figure}

\begin{figure}[h]
    \centering
    \begin{subfigure}{0.48\linewidth}
    \centering
    \includegraphics[width=1.0\linewidth]{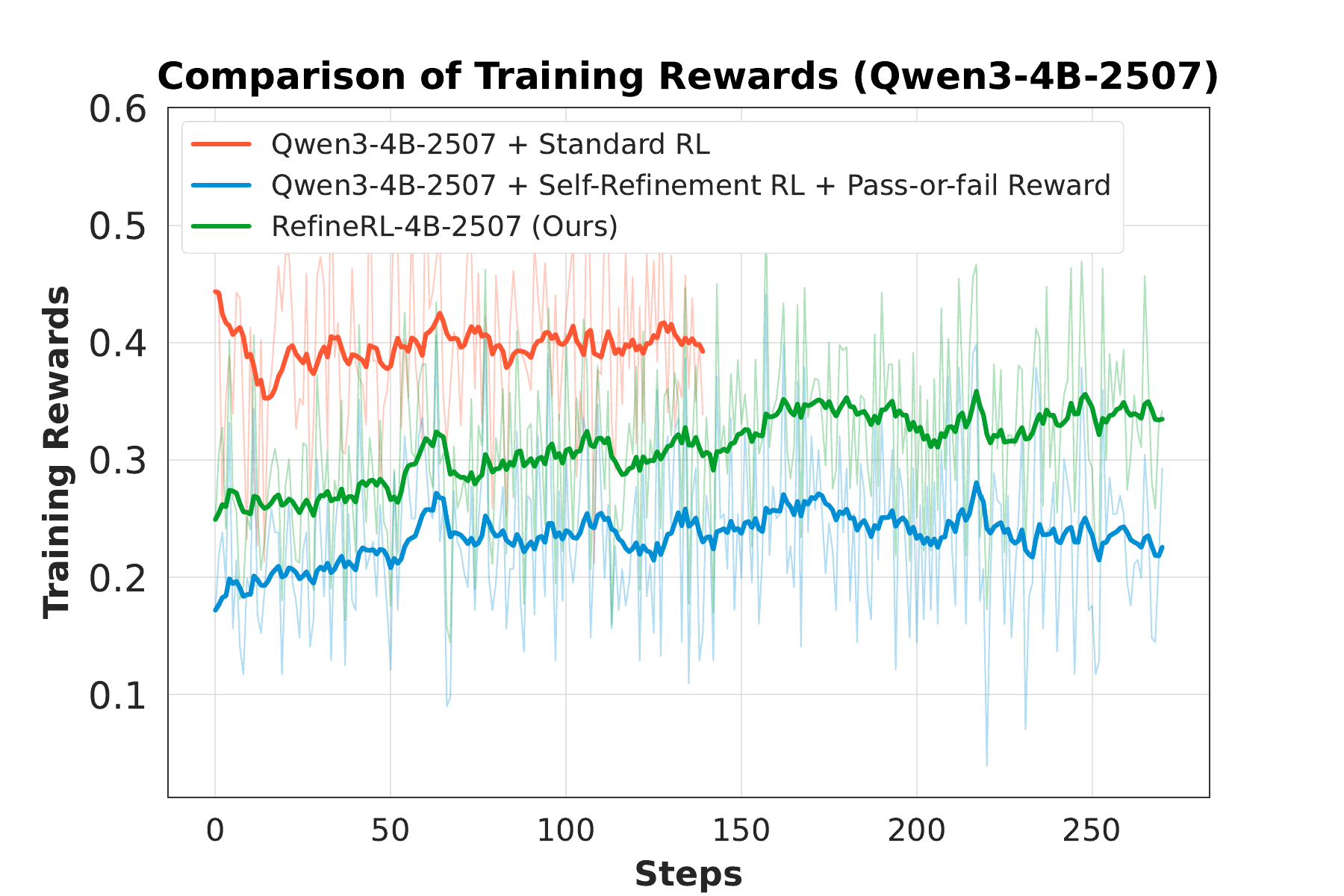}
    \caption{Average reward per step.}
    \end{subfigure}
    \hspace{1em}
    \begin{subfigure}{0.48\linewidth}
    \centering
    \includegraphics[width=1.0\linewidth]{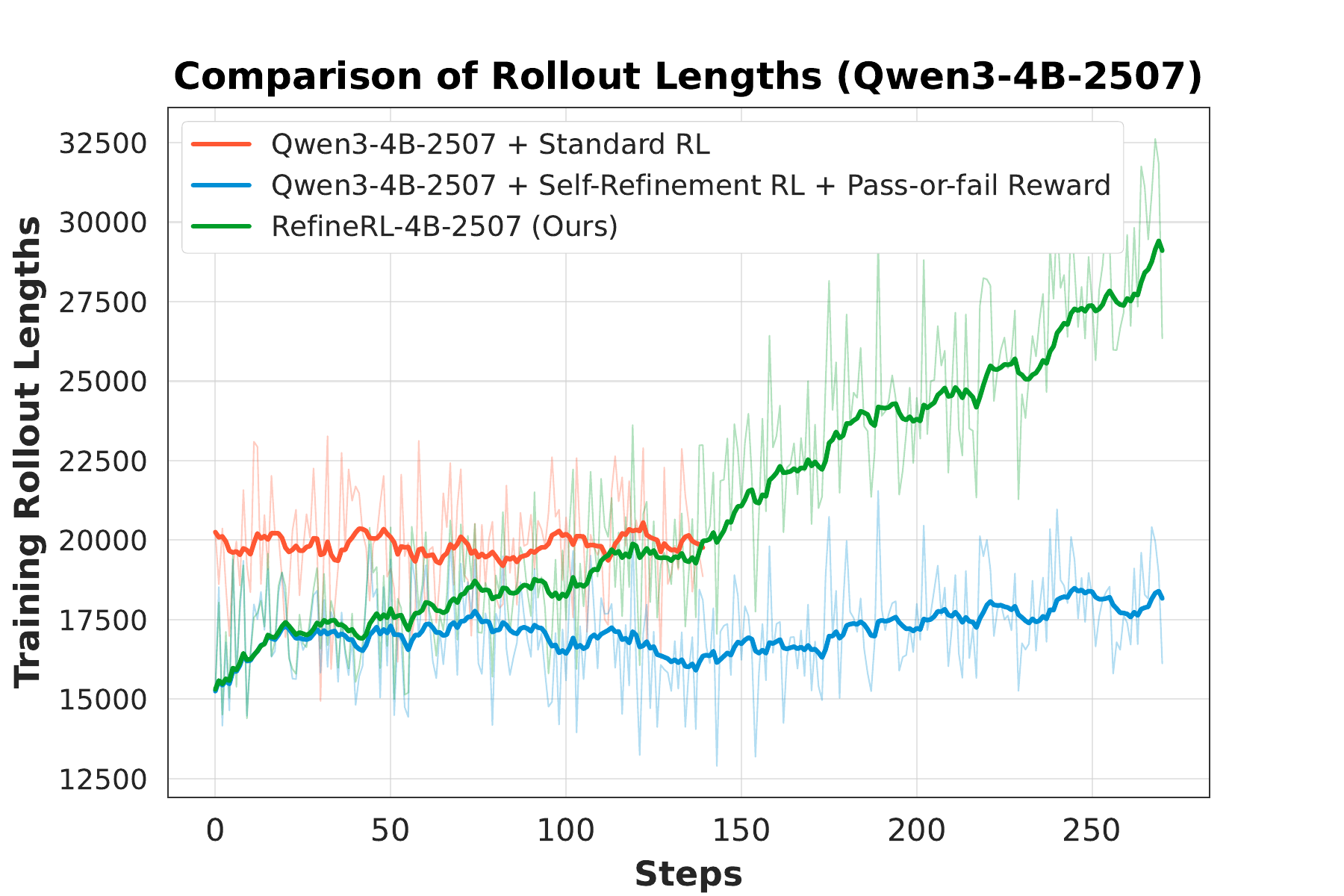}
    \caption{Average rollout length (tokens) per step.}
    \end{subfigure}
    \caption{
    {Training dynamics of Qwen3-4B-2507.}
    Similar to the base model, the 2507 variant demonstrates consistent learning progress in terms of reward maximization and increased reasoning length during the RL process.
    }
    \label{fig:curve:qwen3-4b-2507}
\end{figure}

Figures~\ref{fig:curve:qwen3-4b} and \ref{fig:curve:qwen3-4b-2507} illustrate the evolution of the average reward and the average rollout token length during the training process for \textbf{Qwen3-4B} and \textbf{Qwen3-4B-2507}, respectively.

Across both models, we observe a consistent upward trend in the reward curves of our Self-Refinement RL (\textit{i.e.}, {\ascrl}-4B and {\ascrl}-4B-2507), indicating that the models are effectively learning to solve increasingly complex competitive programming problems and adapting to the squared-incentive reward signal.
Simultaneously, the rollout token lengths of our Self-Refinement RL generally exhibit an increasing trend. This suggests that as the models improve, they learn to generate more comprehensive solution analyses and engage in more extensive self-refinement loops to debug and verify their code before submission, rather than attempting to guess short, likely incorrect solutions.

\subsection{Validation Dynamics}

\begin{figure}[h]
    \centering
    \begin{subfigure}{0.32\linewidth}
    \centering
    \includegraphics[width=1.0\linewidth]{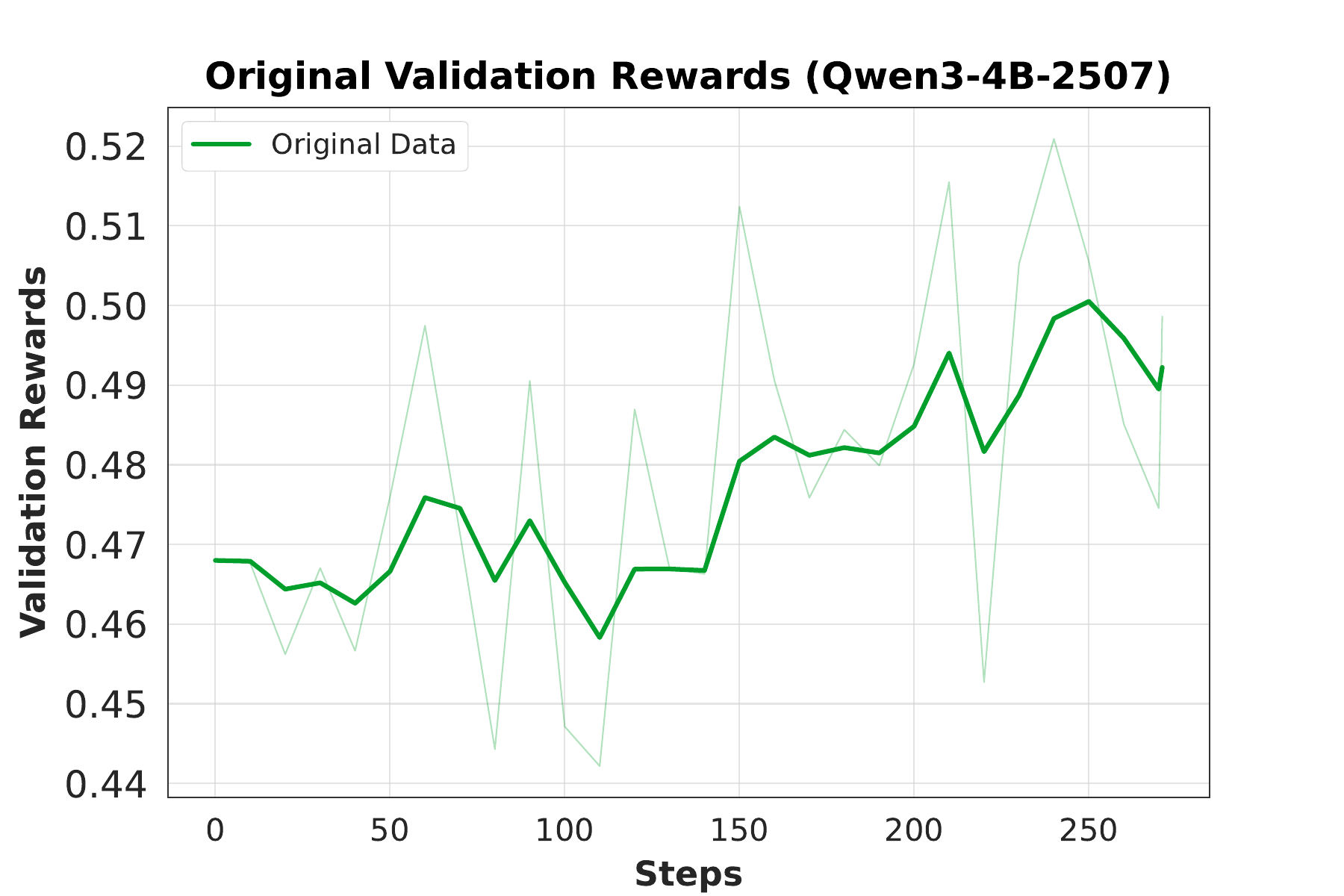}
    \caption{Validation reward per step on the original held-out problems.}
    \end{subfigure}
    \hspace{0.2em}
    \begin{subfigure}{0.32\linewidth}
    \centering
    \includegraphics[width=1.0\linewidth]{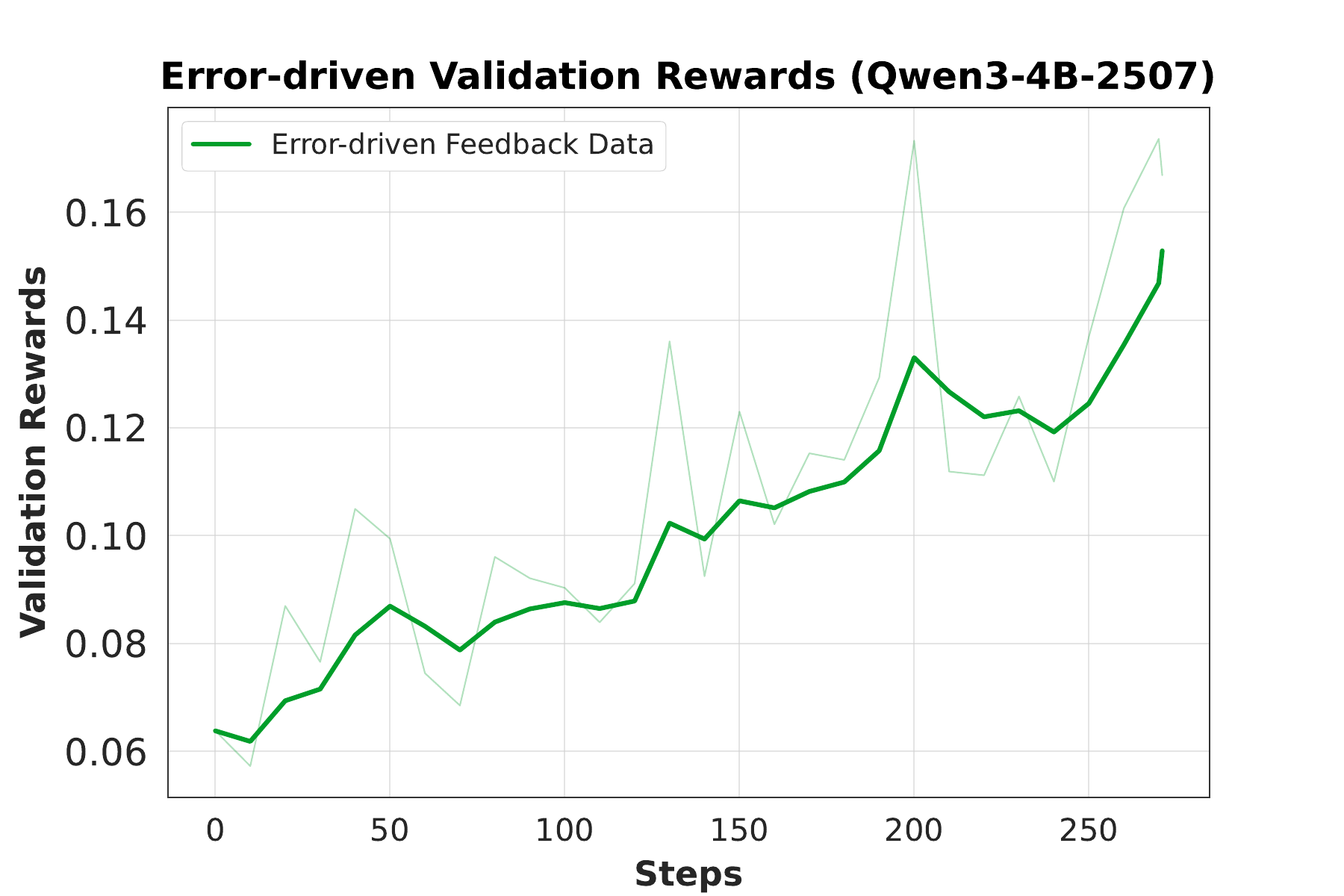}
    \caption{Validation reward on held-out error-driven refinement data.}
    \end{subfigure}
    \hspace{0.2em}
    \begin{subfigure}{0.32\linewidth}
    \centering
    \includegraphics[width=1.0\linewidth]{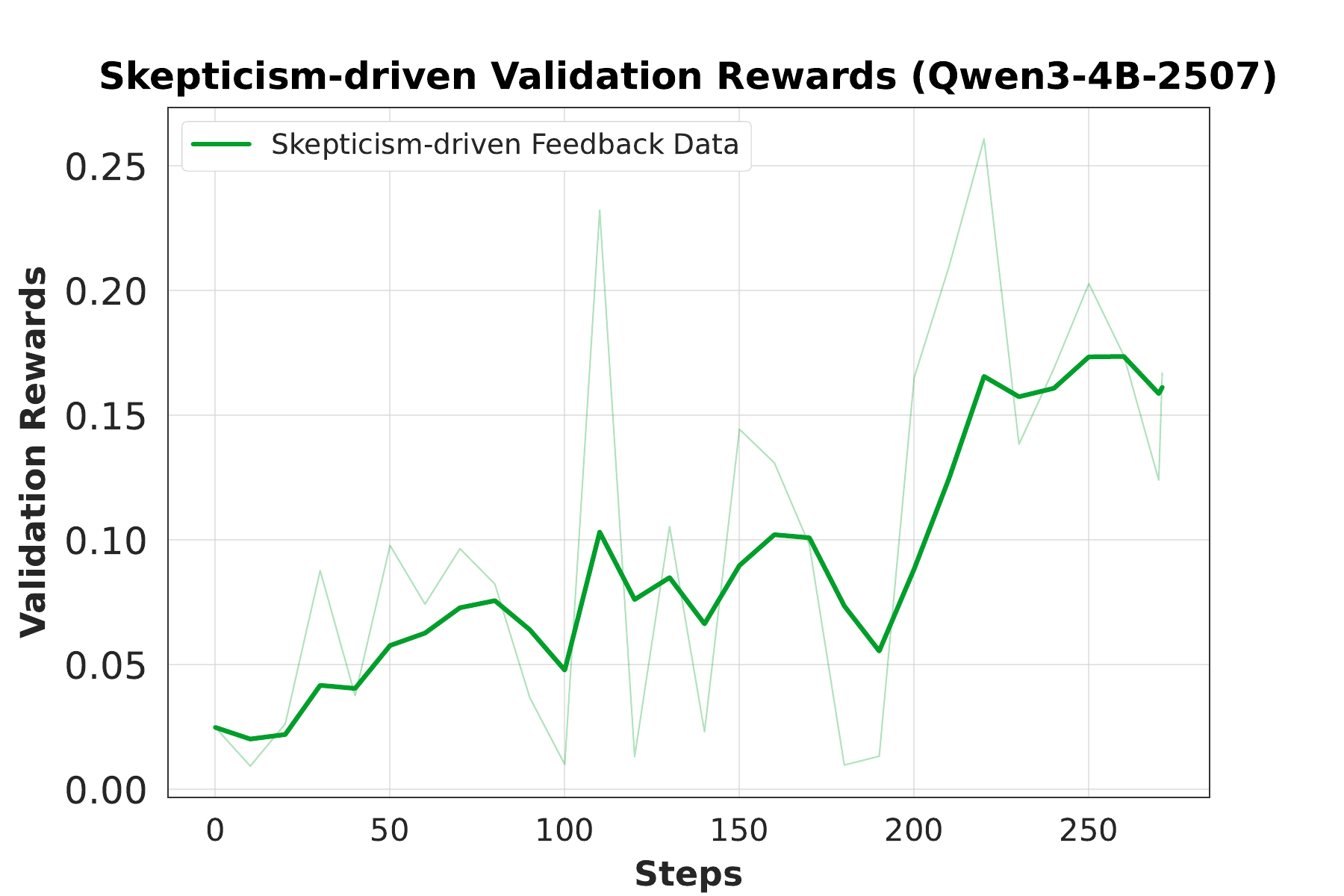}
    \caption{Validation reward on held-out skepticism-driven refinement data.}
    \end{subfigure}
    \caption{
    Validation reward curves across three held-out validation settings of Qwen3-4B-2507 throughout RL training.
    In each subplot, the solid line shows the smoothed trend, while the lighter transparent line shows the raw validation reward.
    These results indicate that the model continues to acquire self-refinement ability during later training stages, rather than merely overfitting to the training trajectories.
    }
    \label{fig:curve-val:qwen3-4b-2507}
\end{figure}

\begin{figure}[h]
    \centering
    \begin{subfigure}{0.32\linewidth}
    \centering
    \includegraphics[width=1.0\linewidth]{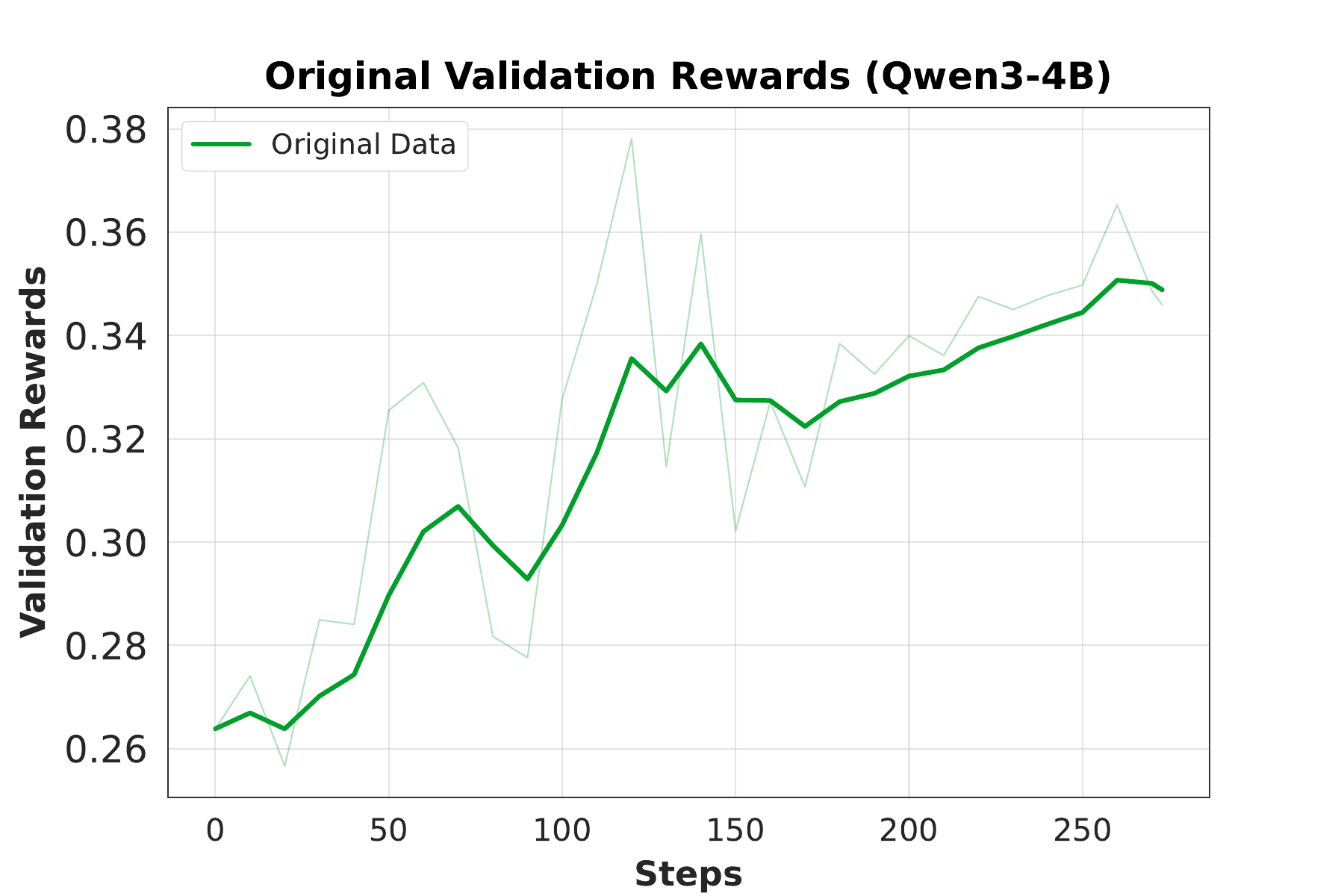}
    \caption{Validation reward per step on the original held-out problems.}
    \end{subfigure}
    \hspace{0.2em}
    \begin{subfigure}{0.32\linewidth}
    \centering
    \includegraphics[width=1.0\linewidth]{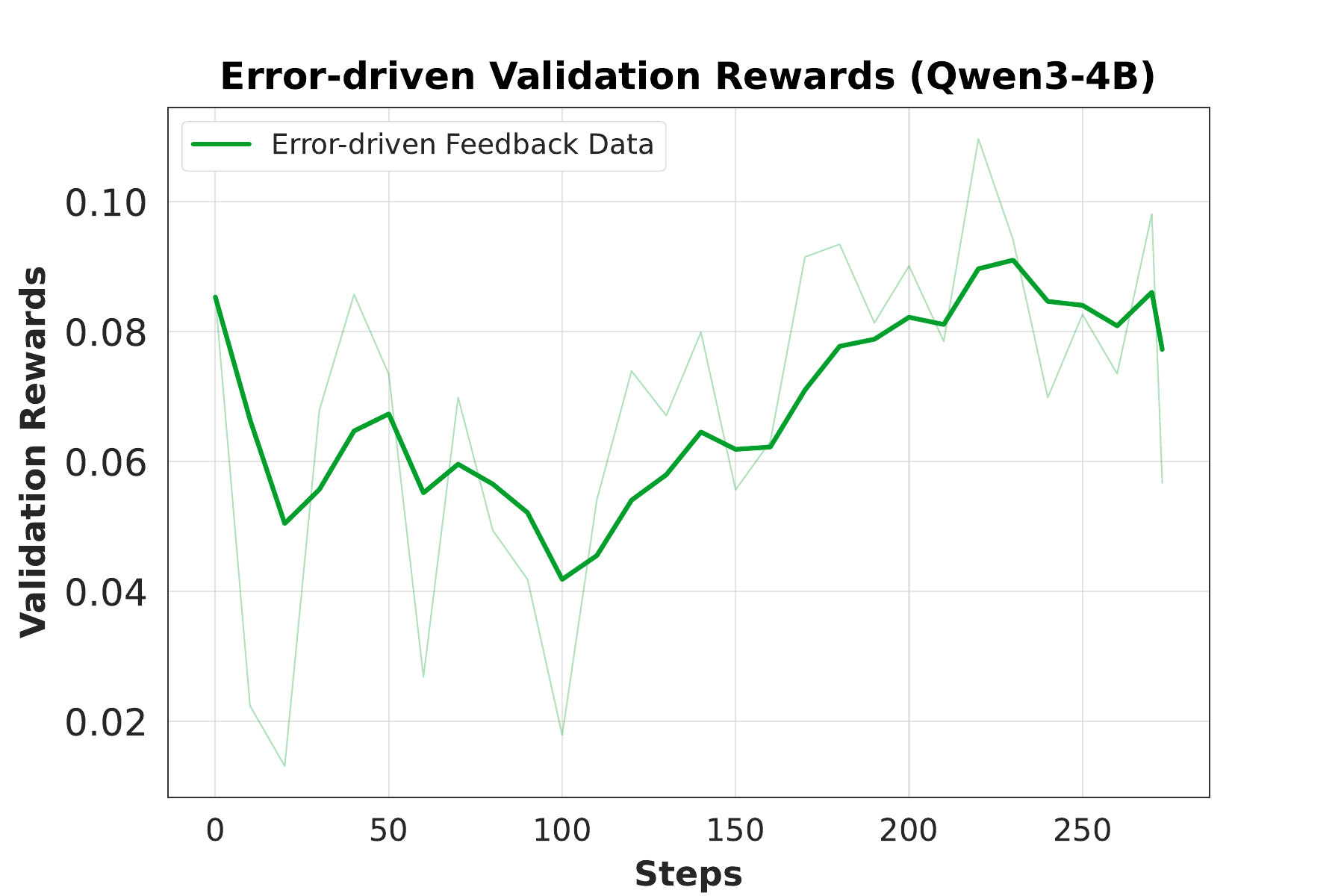}
    \caption{Validation reward on held-out error-driven refinement data.}
    \end{subfigure}
    \hspace{0.2em}
    \begin{subfigure}{0.32\linewidth}
    \centering
    \includegraphics[width=1.0\linewidth]{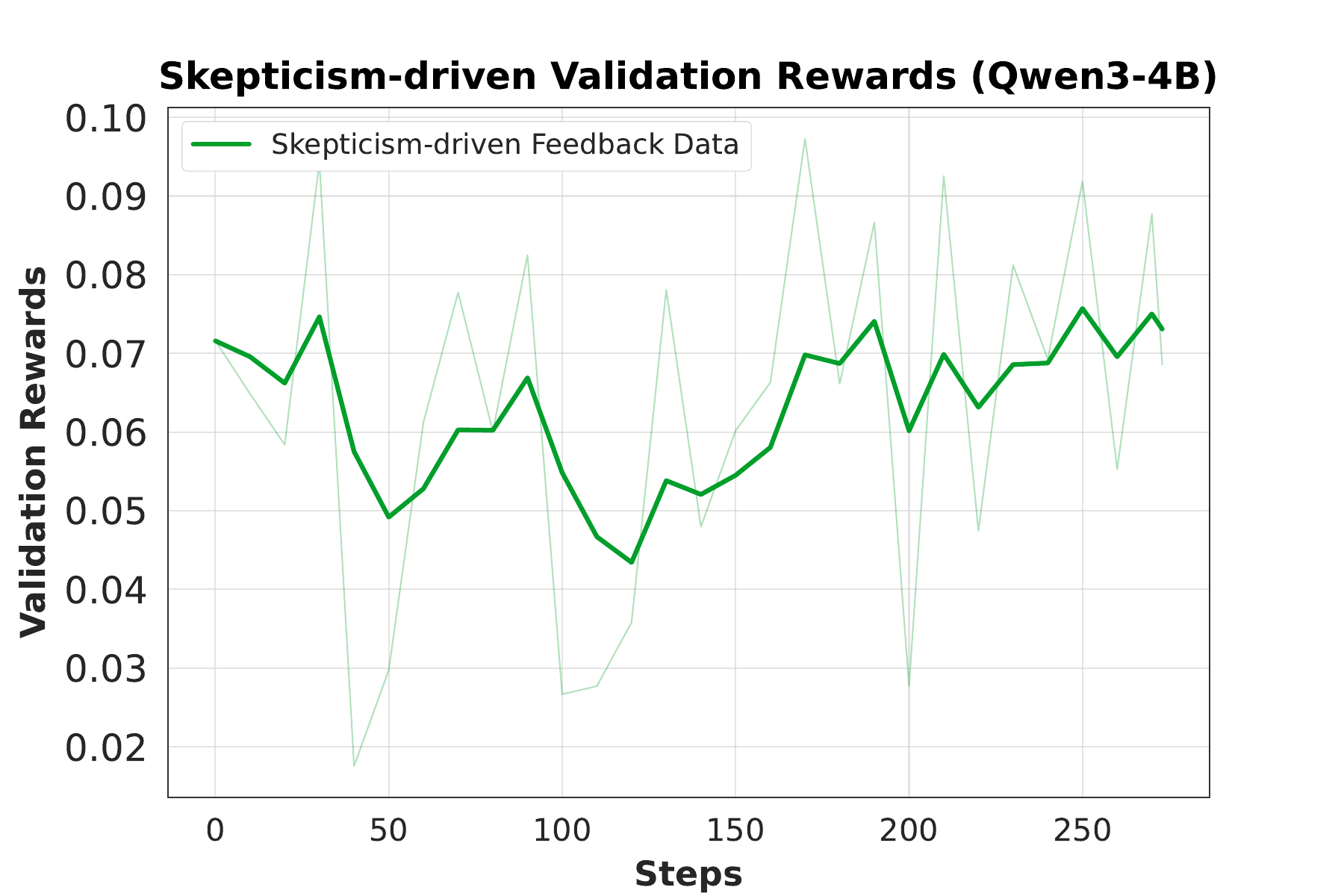}
    \caption{Validation reward on held-out skepticism-driven refinement data.}
    \end{subfigure}
    \caption{
    Validation reward curves across three held-out validation settings of Qwen3-4B throughout RL training.
    In each subplot, the solid line shows the smoothed trend, while the lighter transparent line shows the raw validation reward.
    These results indicate that the model continues to acquire self-refinement ability during later training stages, rather than merely overfitting to the training trajectories.
    }
    \label{fig:curve-val:qwen3-4b}
\end{figure}

We further examine the validation dynamics of Self-Refinement RL to understand whether the model continues to improve during training or merely overfits to the constructed refinement trajectories.
Specifically, we track validation rewards on three held-out settings: original CP problems, error-driven refinement trajectories, and skepticism-driven refinement trajectories.
As shown in Figure~\ref{fig:curve-val:qwen3-4b-2507} and Figure~\ref{fig:curve-val:qwen3-4b}, the validation rewards generally increase across all three settings for both Qwen3-4B-2507 and Qwen3-4B.
This suggests that RL training improves the model's general problem-solving and self-refinement capabilities, rather than only memorizing the training trajectories.

\subsection{Inference-time Strategies on RefinRL-4B}
\label{app:tts-results-refinerl-4b}

\begin{figure}[h]
\centering

\begin{minipage}[t]{0.44\linewidth}
\centering
\captionof{table}{
Performance of inference-time strategies on RefineRL-4B.
Our Skeptical@16 strategy outperforms all compared baselines across all settings.
}
\label{tab:tts-perf:refinerl-4b}
\tiny
\begin{tabular}{l c c c}
\toprule
Model \& Strategy & LCB-v5 & LCB-v6 & USACO \\

\midrule
{\ascrl}-4B (Ours) & 56.40 & 48.79 & 44.75 \\
\quad + Random@16 & 62.95 & 52.68 & 47.72 \\
\quad + LongCoT@16 & 60.14 & 50.32 & 34.93 \\
\quad + RejSamp@16 & 61.49 & 51.75 & 47.72 \\
\quad + Reflexion@16 & 62.39 & 51.86 & 47.94 \\
\rowcolor{green!10}
\quad {+ Skeptical@16 (Ours)} & \textbf{64.07} & \textbf{56.54} & \textbf{50.61} \\

\bottomrule
\end{tabular}
\end{minipage}
\hfill
\begin{minipage}[t]{0.52\linewidth}
\centering
\vspace{0pt}
\includegraphics[width=0.96\linewidth]{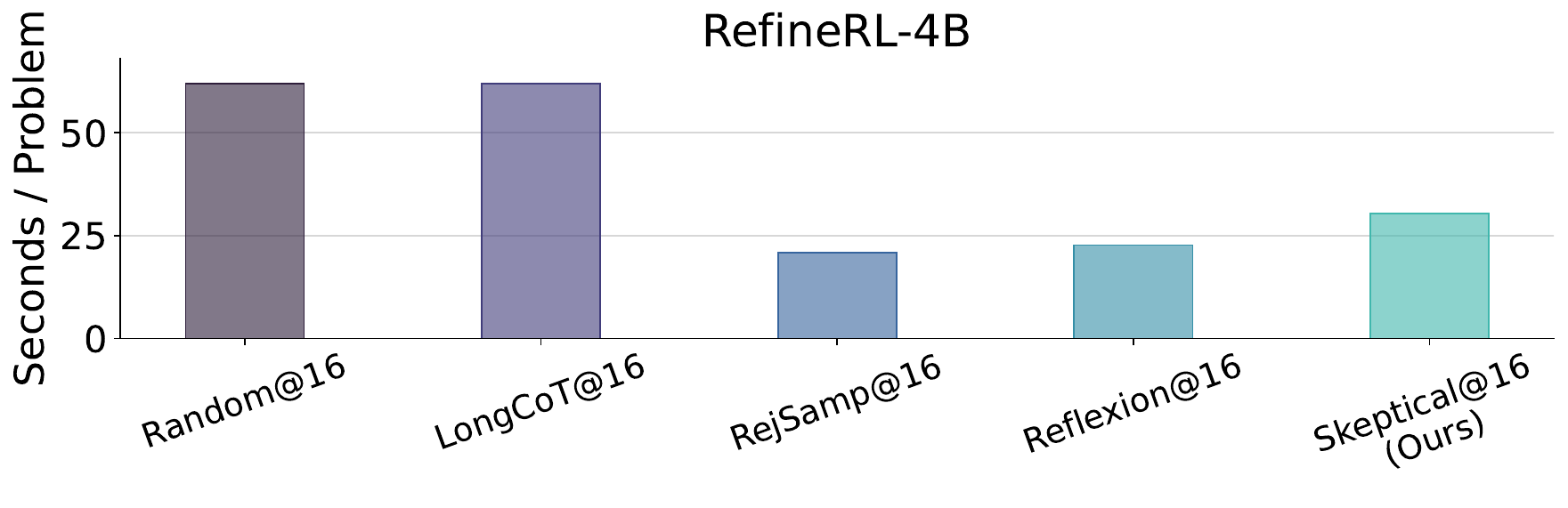}
\caption{
Time cost of inference-time strategies on RefineRL-4B.
Skeptical@16 achieves a speed comparable to other sequential-sampling baselines while being faster than parallel-sampling baselines.
}
\label{fig:tts-time-cost:refinerl-4b}
\end{minipage}

\end{figure}

We report the performance (Table~\ref{tab:tts-perf:refinerl-4b}) and time cost (Figure~\ref{fig:tts-time-cost:refinerl-4b}) of different inference-time strategies on our \ascrl-4B model, following the similar analysis in Section~\ref{sec:exp:results} (see Table~\ref{tab:tts-perf:refinerl-4b-2507} and Figure~\ref{fig:tts-time-cost:refinerl-4b-2507}).
We observe trends similar to those on \ascrl-4B-2507: Skeptical@16 achieves the best overall performance, maintains a speed comparable to those of other sequential baselines, and is substantially faster than parallel-sampling baselines such as Random@16 and LongCoT@16.

\subsection{Case Study on IOI 2024}
\label{app:ioi2024}

We further evaluate {\ascrl} on the International Olympiad in Informatics (IOI) 2024 problems as a challenging case study.
Compared with standard CP benchmarks, IOI problems are substantially more difficult and adopt a subtask-based scoring protocol with multiple submissions allowed for each problem.
We follow the official IOI scoring protocol and report \textbf{Score@50}, where the score of each subtask is maximized over 50 submissions and then aggregated across all problems.
The maximum total score of IOI 2024 is 600.

For Skeptical-Agent, we use the stronger Skeptical@32 setting with $N=32$, $M_1=2$, and $M_2=32$ to accommodate the increased difficulty of IOI problems.
Table~\ref{tab:ioi2024} summarizes the results.

\begin{table}[h]
    \centering
    \tiny
    \setlength{\tabcolsep}{6pt}
    \begin{tabular}{l c}
        \toprule
        \textbf{Model \& Strategy} & \textbf{IOI 2024 Score@50} \\
        \midrule
        Qwen3-235B-A22B-2507 & 148.5 \\
        OCR-Nemotron-1.1-32B & 132.0 \\
        \midrule
        Qwen3-4B-Thinking-2507 & 90.0 \\
        \rowcolor{green!10}
        \quad + Skeptical@32 (Ours) & 109.0 \\
        \rowcolor{green!10}
        {\ascrl}-4B-2507 (Ours) & 109.0 \\
        \rowcolor{green!10}
        \quad + Skeptical@32 (Ours) & \textbf{117.0} \\
        \bottomrule
    \end{tabular}
    \caption{
    Performance on IOI 2024 under the official subtask-level scoring protocol with up to 50 submissions per problem.
    }
    \label{tab:ioi2024}
\end{table}

Self-Refinement RL and Skeptical-Agent independently improve the 4B model from 90.0 to 109.0 points.
Their combination further reaches 117.0, giving a 27.0-point gain over the original model.
This complementary improvement suggests that RL training and inference-time skeptical refinement contribute distinct benefits even on highly challenging IOI problems.

\clearpage
\section{Prompts}
\label{app:prompts}

In this section, we provide the detailed prompt templates used in our Skeptical-Agent.

\subsection{Prompts for Competitive Programming}

Figure~\ref{fig:pmpt:direct-gen} presents the base system prompt used for initializing the problem-solving process and generating the initial solution.
When the agent enters the refinement loop, the feedback it receives depends on the outcome of the local testing.
If the solution fails the local test cases, the agent receives a detailed error report constructed using the template in Figure~\ref{fig:pmpt:p-fail}, which incorporates specific failure reasons (\textit{e.g.}, Wrong Answer, Runtime Error) as detailed in Figure~\ref{fig:pmpt:p-fail:explain}.
Conversely, if the solution passes the local tests, our Skeptical-Agent does not immediately accept it. Instead, it triggers a ``skeptical'' review using the template in Figure~\ref{fig:pmpt:p-pass}, forcing the model to self-verify against potential edge cases, hard-coded cheats, or complexity issues before final submission.

\begin{figure}[h]
\footnotesize
\begin{tcolorbox}[title=Prompt Template for Direct Generation (Python)]
You are a competitive programmer. You will be given a question (problem statement) and will generate a correct python program that matches the specification and passes all tests.

Please first give a solution analysis and then provide the python solution code.

Ensure that when your solution program runs, it reads the input from stdin and writes the output (answer) to stdout. Enclose your code within delimiters as follows:

\verb
```
python

\# YOUR CODE HERE

\verb
```
\\

Question: \{problem statement\}
\\

\# Additional Messages

\{prompt for feedback\}

\end{tcolorbox}
\caption{Standard system prompt used for initial solution generation and subsequent refinement steps.}
\label{fig:pmpt:direct-gen}
\end{figure}

\begin{figure}[h]
\footnotesize
\begin{tcolorbox}[title=Prompt Template for Error-driven Feedback Function $\mathcal P_{\text{Err}}$]
In the previous round, a solution has been proposed but failed to pass the local testing process.
Below provides detailed **Solution Analysis**, **Solution Code**, and **Failure Reason** for this failed solution.
Please consider leveraging all these messages to debug the old solution and further design a better solution.
\\

\#\# Solution Analysis

\verb
```
text

\{analysis\}

\verb
```
\\

\#\# Solution Code

\verb
```
python

\{code\}

\verb
```
\\

\#\# Failure Reason

\{error reason\}
\end{tcolorbox}
\caption{Prompt for error-driven feedback construction.}
\label{fig:pmpt:p-fail}
\end{figure}

\begin{figure}
\footnotesize
\begin{tcolorbox}[title=Error Reasons for Different Evaluation Results]
\begin{tcolorbox}[title=Compile Error (CE)]
The proposed solution code received **Compile Error** during local testing.
\end{tcolorbox}

\begin{tcolorbox}[title=Wrong Answer (WA)]
The proposed solution code received **Wrong Answer Error** during local testing.
The error occurred when the proposed code was running on the following test input:

\verb
```
text

\{test input\}

\verb
```

The expected correct test output is:

\verb
```
text

\{test output\}

\verb
```

However, the output of the proposed solution for this test case is:

\verb
```
text

\{sol test output\}

\verb
```

which is different from the correct output.
\\

**Please DO NOT hard-code this test input/output to your solution code.
This is cheating and would not help you pass the final stress testing evaluation.**
\end{tcolorbox}

\begin{tcolorbox}[title=Runtime Error (RE)]
The proposed solution code received **Runtime Error** during local testing. The error occurred when the proposed code was running on the following test input:

\verb
```
text

\{test input\}

\verb
```

Please pay attention to the details of your code.
\end{tcolorbox}

\begin{tcolorbox}[title=Time Limited Exceed (TLE)]
The proposed solution code received **Time Limited Exceed Error** during local testing. The error occurred when the proposed code was running on the following test input:

\verb
```
text

\{test input\}

\verb
```

Please be aware of the time limit of the problem when solving it.
\end{tcolorbox}

\begin{tcolorbox}[title=Memory Limited Exceed (MLE)]
The proposed solution code received **Memory Limited Exceed Error** during local testing. The error occurred when the proposed code was running on the following test input:

\verb
```
text

\{test input\}

\verb
```

Please be aware of the memory limit of the problem when solving it.
\end{tcolorbox}

\end{tcolorbox}
\caption{Detailed templates for constructing the ``Error Reason'' field based on specific execution errors (WA, TLE, RE, etc.).}
\label{fig:pmpt:p-fail:explain}
\end{figure}

\begin{figure}[h]
\footnotesize
\begin{tcolorbox}[title=Prompt Template for Skepticism-driven Feedback Function $\mathcal P_{\text{Skep}}$]
In the previous round, a solution has been proposed and successfully passed all local test cases. Below provides detailed **Solution Analysis** and **Solution Code** for this solution and an additional **Check List**. Please analyze the proposed solution **step-by-step** following the provided **Check List**.

If the previously proposed solution can pass all conditions in the **Check List**, please explain why and output the solution code. If NOT, please also explain why and leverage all these messages to debug the old solution and further design a better solution to pass the **Check List**.
\\

\#\# Solution Analysis

\verb
```
text

\{analysis\}

\verb
```
\\

\#\# Solution Code

\verb
```
python

\{code\}

\verb
```
\\

\#\#Check List

- Ensure the solution code does NOT hard-code answers for some specific testcases and directly output them to cheat the judging system.

- Ensure the **CORRECTNESS** of the proposed solution by carefully thinking if there are any edge-testcases that can cause the proposed solution code encounter **Wrong Answer Error**.

- Ensure the time and space complexities of the proposed solution are **OPTIMAL** by carefully thinking if there are any edge-testcases that can cause the proposed solution code encounter **Time Limit Exceed Error** or **Memory Limit Exceed Error**.
\end{tcolorbox}
\caption{The skeptical-driven feedback prompt ($\mathcal P_{\text{Skep}}$) triggered when a solution passes local tests (on all public test cases), instructing the agent to self-critique against a checklist.}
\label{fig:pmpt:p-pass}
\end{figure}

\clearpage
\subsection{Prompts for Mathematical Reasoning}

Figure~\ref{fig:pmpt:aime:direct-gen} presents the base prompt used for initial solution generation and subsequent refinement on AIME problems.
Unlike CP tasks, mathematical problems like AIME problems do not provide executable test cases, so we cannot construct error-driven feedback from local testing results.
Therefore, for mathematical reasoning, we enforce refinement solely through skepticism-driven feedback.
Specifically, after each proposed solution, the agent is prompted to re-check the previous reasoning step by step using the template in Figure~\ref{fig:aime:pmpt:p-pass}, and either justify the existing answer or identify errors and propose a corrected solution.

\begin{figure}[h]
\footnotesize
\begin{tcolorbox}[title=Prompt Template for Direct Generation (AIME)]
Solve the following math problem efficiently and clearly. The last line of your response should be of the following format: `Therefore, the final answer is: \$\textbackslash boxed\{ANSWER\}\$. I hope it is correct' (without quotes) where ANSWER is just the final number or expression that solves the problem. Think step by step before answering.
\\

\{problem statement\}
\\

\# Additional Messages

\{prompt for feedback\}
\end{tcolorbox}
\caption{Standard system prompt used for initial solution generation and subsequent refinement steps in solving AIME problems.}
\label{fig:pmpt:aime:direct-gen}
\end{figure}

\begin{figure}[h]
\footnotesize
\begin{tcolorbox}[title=Prompt Template for Skepticism-driven Feedback Function $\mathcal P_{\text{Skep}}$ for AIME]
In the previous round, a solution and its corresponding answer has been proposed. Please analyze the proposed solution **step-by-step** to analyze whether it is indeed correct or not.

If you believe the previously proposed solution is correct, please explain why and output the previous answer. If NOT, please also explain why and leverage all these messages to debug the old solution and propose a better solution.
\\

\#\# Previous Solution

\verb
```text

\{previous solution\}

\verb
```
\end{tcolorbox}
\caption{The skeptical-driven feedback prompt ($\mathcal P_{\text{Skep}}$) triggered every time when the agent tries to perform refinement.}
\label{fig:aime:pmpt:p-pass}
\end{figure}

\end{document}